\documentclass[letterpaper]{article} 
\usepackage{aaai23}  
\usepackage{times}  
\usepackage{helvet}  
\usepackage{courier}  
\usepackage[hyphens]{url}  
\usepackage{graphicx} 
\usepackage{natbib}  
\usepackage{caption} 
\usepackage{booktabs}
\usepackage{multirow}
\usepackage[ruled,vlined]{algorithm2e}
\usepackage{subfigure}

\makeatletter
\DeclareRobustCommand\onedot{\futurelet\@let@token\@onedot}
\def\@onedot{\ifx\@let@token.\else.\null\fi\xspace}

\def\eg{\emph{e.g}\onedot} 
\def\ie{\emph{i.e}\onedot} 
 
\def\etc{\emph{etc}\onedot} \def\vs{\emph{vs}\onedot}

\makeatother

\title{Farsighted Probabilistic Sampling:\\
A General Strategy for Boosting Local Search MaxSAT Solvers}

\author{
    Jiongzhi Zheng$^{1,2}$,
    Kun He$^{1,2}$\thanks{Corresponding author. },
    Jianrong Zhou$^{1,2}$
}
\affiliations{
    $^1$School of Computer Science and Technology, Huazhong University of Science and Technology, China\\
    $^2$Hopcroft Center on Computing Science, Huazhong University of Science and Technology, China\\
    \{jzzheng, brooklet60, jrzhou\}@hust.edu.cn
}

\usepackage{bibentry}

\begin{document}

\maketitle

\begin{abstract}
Local search has been demonstrated as an efficient approach for two practical generalizations of the MaxSAT problem, namely Partial MaxSAT (PMS) and Weighted PMS (WPMS). In this work, we observe that most local search (W)PMS solvers usually flip a single variable per iteration. Such a mechanism may lead to relatively low-quality local optimal solutions, and may limit the diversity of search directions to escape from local optima. To address this issue, we propose a general strategy, called farsighted probabilistic sampling (FPS), to replace the single flipping mechanism so as to boost the local search (W)PMS algorithms. FPS considers the benefit of continuously flipping a pair of variables in order to find higher-quality local optimal solutions. Moreover, FPS proposes an effective approach to escape from local optima by preferring the best to flip among the best sampled single variable and the best sampled variable pair. Extensive experiments demonstrate that our proposed FPS strategy significantly improves the state-of-the-art (W)PMS solvers, and FPS has an excellent generalization capability to various local search MaxSAT solvers. 
\end{abstract}

\section{Introduction}
Maximum Boolean Satisfiability (MaxSAT) is an optimization version of the famous SAT decision problem. Given a propositional formula in the Conjunctive Normal Form (CNF), MaxSAT aims to maximize the number of satisfied clauses. Partial MaxSAT (PMS) is a generalization of MaxSAT, whose clauses are divided into hard and soft. PMS aims to maximize the number of satisfied soft clauses while satisfying all the hard clauses. In a more generalized situation, each soft clause is associated with a positive weight. The resulting problem is called Weighted PMS (WPMS), which aims to maximize the total weight of satisfied soft clauses meanwhile satisfying all the hard clauses. Both PMS and WPMS, denoted as (W)PMS, have many practical applications, such as planning~\cite{Bonet2019}, timetabling~\cite{Demirovic2017}, routing~\cite{Khadilkar2022}, group testing~\cite{Ciampiconi2020}, \etc.

Local search~\cite{Selman1993,Morris1993,Cha1997} is a well-studied category of incomplete algorithms for MaxSAT. One of the most common frameworks of local search MaxSAT algorithms starts from an initial solution, and then flips a variable in each iteration to explore the solution space. Recently, many effective local search strategies have been proposed for (W)PMS, such as approaches for generating high-quality initial solutions~\cite{Cai2017,Cai2020,Zheng2022}, variable selection strategies~\cite{Luo2017,Zheng2022}, and clause weighting schemes~\cite{Cai2014,Luo2017,Lei2018}.

These state-of-the-art local search algorithms~\cite{Cai2014,Luo2017,Lei2018,Cai2020,Zheng2022} have made considerable achievements in solving the (W)PMS. However, they all follow a similar single flipping mechanism to explore the solution space. That is, they only flip a single variable in each iteration. Such a mechanism may cause the algorithm easily fall into a local optimum, \ie, flipping any variable can not improve the current solution. Thus the quality of the local optimal solutions might not be good enough. Moreover, most of these algorithms escape from local optima by a simple random walk strategy~\cite{Cai2014,Lei2018,Cai2020}, \ie, satisfying a randomly selected falsified clause. The high-degree randomness of this strategy may make it hard for these algorithms to find a good direction for the subsequent search.

To handle these issues, we propose a general variable selection strategy, dubbed farsighted probabilistic sampling (FPS), to replace the single flipping mechanism in the local search (W)PMS solvers. FPS employs a two-level technique, that allows the algorithm to look-ahead and considers the benefit of continuously flipping a pair of variables, as well as a probabilistic sampling approach. First, when there exists no single flipping that can improve the current solution (\ie, a local optimum of the single flipping mechanism reaches), FPS tries to look-ahead to find a pair of variables that flipping both can improve the current solution. In this way, the local optimal solutions can be further improved. Second, the two-level technique and the sampling strategy can provide more and better search directions to escape from the local optima. If FPS fails to improve the current solution by flipping a pair of variables, it will choose the best to flip among the best sampled single variable and the best sampled pair of variables.

In the literature, there exist related studies that apply the look-ahead or similar techniques for the SAT problem. Such as the look-ahead technique~\cite{Li2005,Li2007,Wei2008} and an approach using the second level score~\cite{Cai2013-1,Cai2013-2}. To evaluate the benefit of flipping a variable, these two approaches consider the immediate benefit of the flipping, as well as the future benefit that might be obtained after the flipping. However, they still prioritize the immediate benefit of flipping a single variable, and always perform a single flipping per iteration. For example, the look-ahead technique is used to select one among the two best single flipping variables~\cite{Li2007,Wei2008} and the second level score is used only for breaking ties~\cite{Cai2013-1,Cai2013-2}. Thus the look-ahead and second level score techniques play a minor role in selecting the flipping variable. Intuitively, prioritizing the immediate benefit over the future benefit is short-sighted, and may lead to a poor search direction.

There are also some studies proposing multi-flipping local search operators for SAT~\cite{Mali2003} and MaxSAT~\cite{Reisch2020}. But they use an exhaustive method rather than probability sampling in the local search process. Thus they are less efficient when solving large instances. In contrast, our proposed FPS algorithm combines the advantages of look-ahead and probability sampling. Thus FPS is more effective and efficient for solving the (W)PMS.

To evaluate the performance of FPS, we apply FPS to some of the state-of-the-art local search (W)PMS algorithms, including BandMaxSAT~\cite{Zheng2022}, SATLike3.0~\cite{Cai2020} (the newest extension of SATLike~\cite{Lei2018}), CCEHC~\cite{Luo2017}, and Dist~\cite{Cai2014}. Details of how to apply FPS to these algorithms are described in Section \ref{sec-usage}. The results show that these algorithms can all be improved significantly by FPS, demonstrating its excellent generalization capability. Moreover, we make a further comparison with some of the state-of-the-art SAT-based (W)PMS solvers, including SATLike-c~\cite{Lei2021}, TT-Open-WBO-Inc~\cite{Nadel2019}, and Loandra~\cite{Berg2019}. We apply FPS to improve the local search component in SATLike-c, the resulting solver also outperforms these SAT-based (W)PMS solvers.

The main contributions of this work are as follows:

\begin{itemize}
\item We propose a general farsighted probabilistic sampling (FPS) strategy for boosting local search MaxSAT solvers. Extensive experiments demonstrate that FPS significantly improves the state-of-the-art local search (W)PMS solvers as well as one of the state-of-the-art SAT-based solvers, SATLike-c, which won three among all the four incomplete tracks in MaxSAT Evaluation 2021.
\item FPS can improve the local optimal solutions of the single flipping mechanism by considering the benefit of continuously flipping a pair of variables. FPS escapes from local optima by preferring the best to flip among the best sampled single variable and the best sampled pair of variables, which can provide high-quality search directions to escape from local optima.
\item Our method suggests an efficient and effective way to apply the look-ahead strategy and the multiple flipping mechanism to boost local search MaxSAT solvers, showing great potential for these approaches for MaxSAT.
\end{itemize}

\section{Preliminary}
In this section, we first present formal definitions of the studied problems and some important concepts. Then we summarize the general framework of the state-of-the-art local search (W)PMS algorithms based on the single flipping mechanism, which can help understand how FPS can be applied to boost them (described in Section \ref{sec-usage}).

\subsection{Definitions on Problems and Concepts}
\label{sec-def}
Given a set of Boolean variables $\{x_1,...,x_n\}$, a literal is either a variable $x_i$ itself or its negation $\lnot x_i$, a clause $c$ is a disjunction of literals, and a conjunctive normal form (CNF) formula $\mathcal{F}$ is a conjunction of clauses. A complete assignment $A$ is a mapping that assigns to each variable either 1 (true) or 0 (false). A literal $x_i$ (resp. $\lnot x_j$) is true if $x_i = 1$ (resp. $x_j = 0$). A clause is satisfied if it has at least one true literal, and falsified otherwise.

Given a CNF formula, SAT aims to determine whether there is an assignment that satisfies all the clauses in the formula, and MaxSAT aims to find an assignment that maximizes the number of satisfied clauses. The PMS, for which the clauses are divided into hard and soft, aims to find an assignment that satisfies all the hard clauses and satisfies as many soft clauses as possible. The WPMS, for which the soft clauses are associated with positive weights, aims to find an assignment that satisfies all hard clauses and maximizes the total weight of the satisfied soft clauses.

For a (W)PMS instance $\mathcal{F}$, an assignment $A$ is regarded as feasible if it satisfies all the hard clauses in $\mathcal{F}$, and the cost of a feasible assignment $A$, denoted by $cost(A)$, is defined to be the number (or total weight) of the falsified soft clauses. For convenience, the cost of any infeasible assignment is set to $+\infty$. The flipping operator in local search algorithms for MaxSAT on a variable is to change its Boolean value. Recent effective local search (W)PMS algorithms use the clause weighting strategy that associates dynamic weights to both hard and soft clauses to guide the search direction. Some of them, such as BandMaxSAT and SATLike(3.0), use a single scoring function $score(x)$ to represent the increase of the total dynamic weight of the satisfied hard and soft clauses caused by flipping $x$. Others such as CCEHC and Dist use $hscore(x)$ (resp. $sscore(x)$) to represent the increase of the total dynamic weight of the satisfied hard (resp. soft) clauses caused by flipping $x$.

\subsection{General Single Flipping Local Search}
\label{sec-single}
The general flow of local search (W)PMS algorithms based on the single flipping mechanism is shown in Algorithm~\ref{alg_single}. We mainly focus on the variable selection process in each iteration (lines 5-9), which can be divided into two cases. In the first case (lines 5-6), the current assignment $A$ is not a local optimum, \ie, $GoodVars \neq \emptyset$. Different algorithms define different $GoodVars$. For example, BandMaxSAT and SATLike(3.0) define $GoodVars$ as $\{x|score(x) > 0\}$, Dist and CCEHC define $GoodVars$ as $\{x|hscore(x) > 0 \vee (hscore(x) = 0 \wedge sscore(x) > 0)\}$. CCEHC further uses the configuration checking strategy~\cite{Cai2011} to refine $GoodVars$. In general, flipping a variable in $GoodVars$ results in a better solution than the current one. The algorithms use a greedy~\cite{Luo2017} or sampling strategy~\cite{Lei2018,Cai2020} to select a variable to be flipped in this case. Anyhow, the variable selection strategies of these algorithms in this case are reasonable, since the current solution can always be improved.

In the second case (lines 7-9), the algorithms fall into a local optimum. A common strategy is to select a falsified clause $c$ (usually with a bias to hard clauses) first (line 8), and then select a variable $v$ from $c$ (line 9). The Dist, CCEHC, and SATLike(3.0) algorithms use the simple random walk strategy to select the clause to be satisfied (\ie, $c$) randomly. The recently proposed BandMaxSAT algorithm uses its multi-armed bandit model to select the clause $c$ smartly, which can help the algorithm find a good search direction. After determining $c$, the algorithms usually select $v$ greedily according to their scoring functions. Besides, before using the random walk strategy, CCEHC tries to greedily (according to its scoring functions) select the variable to be flipped that satisfies the configuration checking condition~\cite{Luo2017} in all the falsified clauses.

\begin{algorithm}[t]
\caption{General Single Flipping Local Search}
\label{alg_single}
\LinesNumbered 
\KwIn{A (W)PMS instance $\mathcal{F}$, cut-off time \textit{cutoff}}
\KwOut{A feasible assignment $A$ of $\mathcal{F}$, or \textit{no feasible assignment found}}
$A :=$ an initial assignment; $A^* := A$\;
\While{\textit{running time} $<$ \textit{cutoff}}{
\If{$A$ is feasible $\&$ \textit{cost}($A$) $<$ \textit{cost}($A^*$)}{
$A^* := A$\;
}
\eIf{$GoodVars \neq \emptyset$}{
$v :=$ a variable selected from $GoodVars$\;
}{
$c :=$ a selected falsified clause\;
$v :=$ a variable selected from $c$\;}
$A := A$ with $v$ flipped\;
}
\lIf{$A^*$ is feasible}{\textbf{return} $A^*$}
\lElse{\textbf{return} \textit{no feasible assignment found}}
\end{algorithm}

\section{Methodology}
\label{Sec-FPS}
We propose a general farsighted probabilistic sampling (FPS) variable selection strategy to replace or improve the single flipping mechanism that is widely used in local search (W)PMS solvers. This section first introduces the main process of FPS, and then introduces how to use FPS to improve the state-of-the-art local search (W)PMS solvers.

\subsection{The Proposed FPS Strategy}
As described in Section \ref{sec-single}, 
the variable selection strategies (lines 5-6 in Algorithm \ref{alg_single}) based on the single flipping mechanism are reasonable when local optima are not reached, since the current solution can always be improved. However, when the algorithms fall into a local optimum, they stop improving the current solution and allow the search to get a worse solution to escape from the local optimum. We argue that such a strategy may make the algorithms miss better solutions, and we propose FPS to handle this issue by using a two-level look-ahead technique. When a local optimum for the single flipping mechanism is reached, FPS first samples some first-level variables, and then tries to look-ahead from these variables to see whether flipping a pair of variables can improve the current solution. In this way, a local optimum for the single flipping mechanism might not be a local optimum for FPS. Moreover, FPS escapes from its local optima by selecting the best to flip among the best sampled first-level variable and the best sampled pair of variables, which can provide a better search direction than the widely-used random walk local optima escaping strategy.

\begin{algorithm}[!t]
\caption{General Local Search based on FPS}
\label{alg_FPS}
\LinesNumbered 
\KwIn{A (W)PMS instance $\mathcal{F}$, cut-off time \textit{cutoff}, number of sampled clauses $sc\_num$, BMS parameter $sv\_num$}
\KwOut{A feasible assignment $A$ of $\mathcal{F}$, or \textit{no feasible assignment found}}
$A :=$ an initial assignment; $A^* := A$\;
\While{\textit{running time} $<$ \textit{cutoff}}{
\If{$A$ is feasible $\&$ \textit{cost}($A$) $<$ \textit{cost}($A^*$)}{
$A^* := A$\;
}
\eIf{$GoodVars := \{x|score(x) > 0\} \neq \emptyset$}{
$v :=$ a variable selected from $GoodVars$\;
$Vars := \{v\}$\;
}
{
$SC :=$ the set of $sc\_num$ randomly selected falsified clauses (with a bias to hard ones)\;
The set of first-level variables $FV := \emptyset$\;
\For{$i := 1$ to $sc\_num$}{
$v :=$ a random variable in $SC_i$\;
\lIf{$v \notin FV$}{$FV := FV \cup \{v\}$}
}
$v_1 := \mathop{\arg\max}\limits_{v \in FV}{score(v)}$\; 
$s_1 := score(v_1)$; $s_2 := -\infty$\;
\For{$i := 1$ to $|FV|$}{
$score' := score(FV_i)$\; 
Performing a pseudo flipping for $FV_i$\;
\If {$GoodVars' := \{x|score(x) > 0\} \neq \emptyset$}{
$SV_i := $ a second-level variable in $GoodVars'$ picked by BMS with parameter $sv\_num$\;
\If{$score' + score(SV_i) > 0$}{$Vars := \{FV_i, SV_i\}$\;
$s_2 := score' + score(SV_i)$\;
\textbf{break}\;}
\If{$score' + score(SV_i) > s_2$}{$v_2^1 := FV_i$; $v_2^2 := SV_i$\;
$s_2 := score' + score(SV_i)$\;}}
}
\lIf{$s_1 > s_2$}{$Vars := \{v_1\}$}
\lElse{$Vars := \{v_2^1, v_2^2\}$}
}
$A := A$ with the variables in $Vars$ flipped\;
}
\lIf{$A^*$ is feasible}{\textbf{return} $A^*$}
\lElse{\textbf{return} \textit{no feasible assignment found}}
\end{algorithm}

The procedure of a general local search algorithm based on FPS is shown in Algorithm \ref{alg_FPS}. Note that we use the single scoring function $score(\cdot)$ to depict the procedure, which is easier to understand. For the algorithms that the scoring functions regarding hard and soft clauses are calculated independently, we just need to replace $score(\cdot)$ with $hscore(\cdot)$ and $sscore(\cdot)$ accordingly.

When $GoodVars \neq \emptyset$, FPS also selects the variable to be flipped from $GoodVars$ (lines 5-7) as Algorithm \ref{alg_single} does. When $GoodVars = \emptyset$, FPS first samples a set of first-level variables (lines 9-13), denoted as $FV$, and then look-ahead from each first-level variable (lines 16-27). To determine $FV$, the algorithm first samples $sc\_num$ (10 by default) falsified hard clauses, if any; otherwise, samples $sc\_num$ falsified soft clauses, and then randomly samples a first-level variable in each sampled clause. Such a sampling strategy for selecting $FV$ is effective in FPS. On the one hand, look-ahead from all the variables in all the falsified clauses is time-consuming. Thus the sampling strategy can improve the efficiency. On the other hand, sampling first-level variables from multiple clauses can provide diverse search directions, \ie, satisfying different clauses.

After sampling $FV$, the first-level variable with the highest $score$ is recorded as $v_1$, whose $score$ is recorded as $s_1$ (lines 14-15). Then, FPS tries to perform a pseudo flipping for each first-level variable $FV_i$ (line 18). Note that the pseudo flipping operator will not change the current solution and maintained information, just to look-ahead to determine $GoodVars'$ (line 19), which records the second-level variables with a positive $score$ after flipping $FV_i$ (\ie, the $score$ of each second-level variable is the one computed after performing the pseudo flipping of $FV_i$). Such a pseudo flipping operator can avoid the redundant flipping operator for restoring the current solution $A$ after each look-ahead process.

If $GoodVars' = \emptyset$, which means flipping both $FV_i$ and any second-level variable in $GoodVars'$ will not be better than only flipping $FV_i$. In this case, look-ahead from $FV_i$ can not gain further benefits, and the algorithm will continue to look-ahead from the next first-level variable. If $GoodVars' \neq \emptyset$, flipping both $FV_i$ and some second-level variable $SV_i$ in $GoodVars'$ might improve the current local optimal solution. FPS selects $SV_i$ by a probabilistic sampling strategy called Best from Multiple Selections (BMS)~\cite{Cai2015}, which chooses $sv\_num$ random variables from $GoodVars'$ and returns the one with the highest $score$ (line 20). Once a pair of variables $(FV_i, SV_i)$ that flipping both can improve the current local optimum is found, FPS uses an early-stop strategy (lines 21-24) to terminate the traversing of $FV$. The best sampled pair of variables are recorded as $(v_2^1, v_2^2)$, and the total $score$ of flipping them is recorded as $s_2$ (lines 26-27). Finally, if FPS fails to improve the current solution, it selects the best to flip among $v_1$ and $(v_2^1, v_2^2)$ according to the benefits of flipping them, \ie, $s_1$ and $s_2$ (lines 28-29).

\subsection{Applying FPS to Various Local Searches}
\label{sec-usage}
As shown in Algorithm \ref{alg_FPS}, FPS actually provides a two-level look-ahead strategy for selecting the variables to be flipped when the local optima for the single flipping mechanism are reached. Therefore, a simple application of FPS to a local search (W)PMS algorithm is to replace its local optima escaping strategy (\eg, lines 7-9 in Algorithm \ref{alg_single}) with that in FPS (\ie, lines 8-29 in Algorithm \ref{alg_FPS}). In this work, we use this method to apply FPS to SATLike3.0~\cite{Cai2020}, CCEHC~\cite{Luo2017}, and Dist~\cite{Cai2014}. Among them, SATLike3.0, the newest extension of the famous SATLike~\cite{Lei2018} algorithm, is the best-performing one and also a typical algorithm based on the single flipping mechanism and the simple random walk local optima escaping strategy. Therefore, we choose SATLike3.0 as the core baseline algorithm. The resulting algorithm of applying FPS to SATLike3.0 is called MaxFPS (available at https://github.com/JHL-HUST/FPS). CCEHC and Dist are two representative algorithms for WPMS and PMS, respectively. We use CCEHC-FPS (resp. Dist-FPS) to represent the algorithm of applying FPS to CCEHC (resp. Dist).

We further apply FPS to improve the recently proposed local search (W)PMS algorithm, BandMaxSAT~\cite{Zheng2022}. Since BandMaxSAT mainly uses its bandit model to select the search direction to escape from the local optima, we can not keep the core features of BandMaxSAT by simply replacing its local optima escaping strategy with that in FPS. 
Therefore, we propose another way to apply FPS to BandMaxSAT. That is, the first-level variables $FV$ in FPS are sampled from the clause selected by the bandit model in BandMaxSAT. The resulting solver is called BandMaxSAT-FPS. The application of FPS in BandMaxSAT-FPS indicates that one can design appropriate methods to determine $FV$ to use FPS flexibly.



\section{Experiments}
For experiments, we first analyze the influence of parameters $sc\_num$ and $sv\_num$ on the performance of FPS. Then we compare the algorithms improved by FPS with the baselines, \ie, MaxFPS \vs SATLike3.0, CCEHC-FPS \vs CCEHC, Dist-FPS \vs Dist, and BandMaxSAT-FPS \vs BandMaxSAT, to evaluate the performance of FPS. We further replace the local search component in SATLike-c~\cite{Lei2021}, \ie, SATLike3.0, with MaxFPS, and compare the resulting solver MaxFPS-c with some of the state-of-the-art SAT-based (W)PMS solvers, including SATLike-c, TT-Open-WBO-Inc~\cite{Nadel2019}, and Loandra~\cite{Berg2019}. Finally, we do ablation studies for clarity and in-depth analysis. 

Note that since the typical SATLike3.0 algorithm based on the single flipping mechanism is our core baseline algorithm, we use its improved algorithm MaxFPS to analyze the parameters and do ablation studies.

\subsection{Experimental Setup}
For the (W)PMS experiments, we select all the instances from the incomplete track of the last four MaxSAT Evaluations (MSEs), \ie, MSE2018 to MSE2021, for comparison and analysis. We use (W)PMS\_$y$ to represent the (W)PMS benchmarks in MSE of year $y$. All the algorithms in the experiments were implemented in C++ and run on a server using an Intel® Xeon® E5-2650 v3 2.30 GHz CPU and 256 GB RAM, running Ubuntu 16.04 Linux operation system. Each (W)PMS instance is solved once (as the baselines and MSE do) by each algorithm with two time limits, 60 and 300 seconds, which are consistent with the settings in MSEs. The random seed for each algorithm is set to 1 as the baseline algorithms do. Due to the limited space, this section only presents the results within 300 seconds of time limit. See results within 60 seconds of time limit in Appendix. 

\subsection{Parameter Study}
\label{sec-para}
The parameters in FPS include $sc\_num$, the number of sampled clauses, and $sv\_num$, the BMS parameter for sampling the second-level variables. We compare 12 different settings of $sc\_num \in \{5,10,20,50\}$ and $sv\_num \in \{20,50,100\}$ (selected according to our experience) based on MaxFPS on all the (W)PMS instances from the incomplete track of MSE2017. We use the scoring function in MSEs to calculate a score for each algorithm per instance, which equals to zero if the output solution is infeasible, otherwise to the cost (see Section \ref{sec-def}) of the best-known solution plus 1 divided by the cost of the output solution plus 1.

\begin{figure}[t]
\centering
\includegraphics[width=1.0\columnwidth]{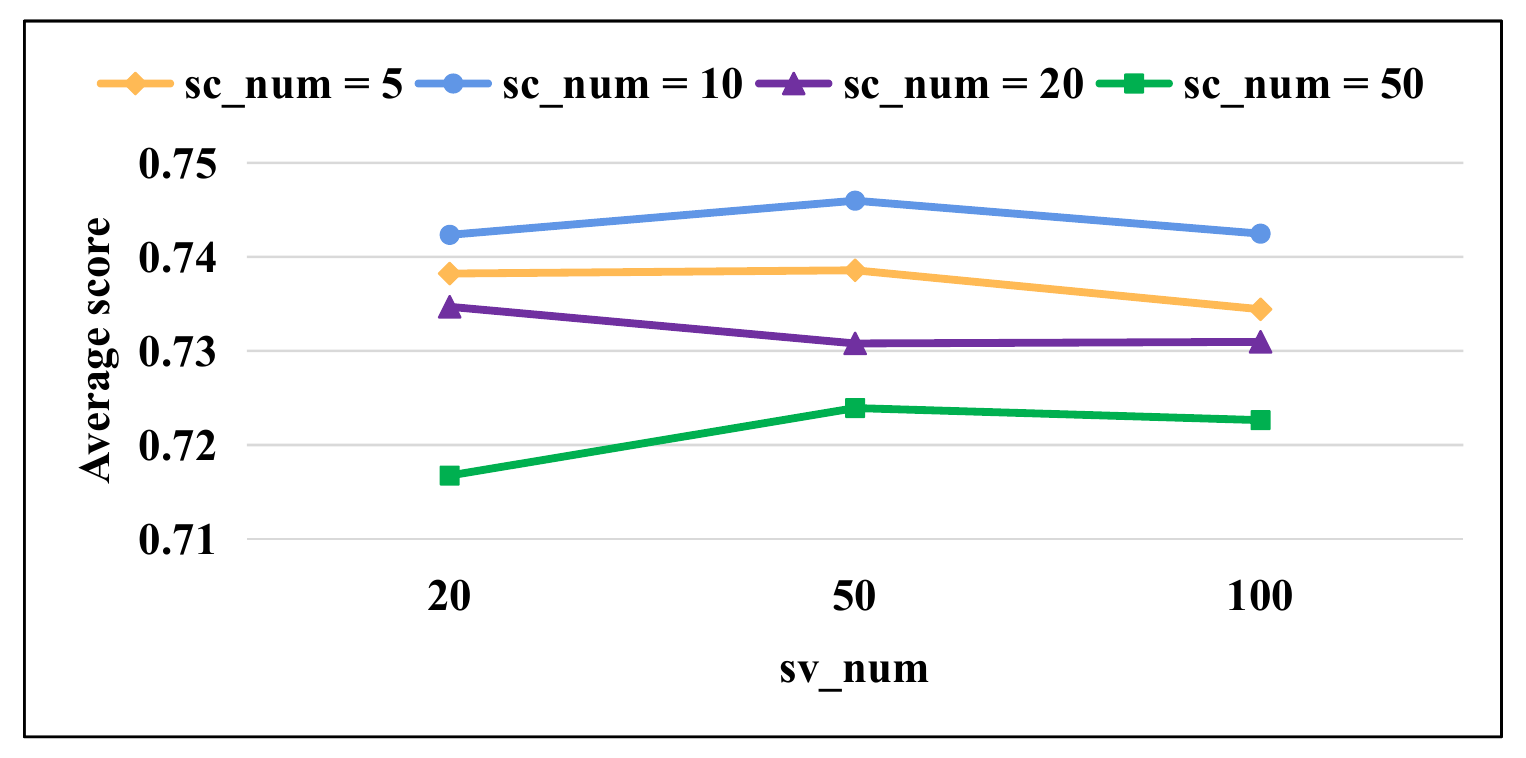}
\caption{Comparison on MaxFPS with different settings of $sc\_num$ and $sv\_num$.}
\label{fig_para}
\end{figure}

Figure \ref{fig_para} shows the comparison results of MaxFPS with different parameters. The results are expressed by the average score of all the tested instances in MSE2017. From the results, we can see that assigning moderate values to both $sc\_num$ and $sv\_num$ results in a good performance. Actually, the larger the value of $sc\_num$ or $sv\_num$, the higher the quality of the local optima for MaxFPS, and the lower the algorithm efficiency. Therefore, moderate values of the parameters can well balance the search ability and algorithm efficiency. The results also demonstrate that the sampling strategies used in FPS are effective. We select the best among the 12 tested settings, \ie, $sc\_num = 10$ and $sv\_num = 50$, as the default parameters in all the algorithms improved by FPS in our experiments.

\subsection{Evaluation on FPS}
This subsection first presents a comprehensive comparison between MaxFPS and SATLike3.0 to evaluate the performance of FPS, and then compares the other algorithms improved by FPS and the corresponding baselines to evaluate the generalization capability of FPS.

\begin{table}[t]
\footnotesize
\centering
\begin{tabular}{lrrrrrr} \bottomrule
\multicolumn{1}{l}{\multirow{2}{*}{Benchmark}} & \multirow{2}{*}{\#inst.} &   \multicolumn{2}{c}{MaxFPS} &  & \multicolumn{2}{c}{SATLike3.0} \\ \cline{3-4} \cline{6-7}
\multicolumn{1}{c}{}                           &                            & \#win.            & time       &  & \#win.       & time        \\ \hline
PMS\_2018                                      & 153                        & \textbf{111}      & 60.68       &  & 61           & 83.43       \\
PMS\_2019                                      & 299                        & \textbf{212}      & 49.05      &  & 142          & 56.64       \\
PMS\_2020                                      & 262                        & \textbf{189}      & 41.09       &  & 112          & 62.14        \\
PMS\_2021                                      & 155                        & \textbf{108}      & 47.12       &  & 64           & 51.00 
\\
WPMS\_2018                                     & 172                        & \textbf{115}      & 78.90       &  & 64           & 84.37        \\
WPMS\_2019                                     & 297                        & \textbf{222}      & 94.71       &  & 100          & 80.38        \\
WPMS\_2020                                     & 253                        & \textbf{183}      & 92.49       &  & 86           & 72.82        \\
WPMS\_2021                                     & 151                        & \textbf{84}       & 104.92       &  & 64           & 98.31        \\
\toprule
\end{tabular}
\caption{Comparison of MaxFPS and SATLike3.0.}
\label{table_MaxFPS}
\end{table}
\begin{table}[t]
\footnotesize
\centering
\scalebox{0.83}{\begin{tabular}{lrrrrrr} \bottomrule
\multirow{2}{*}{Benchmark}         & \multirow{2}{*}{\#inst.} & \multicolumn{2}{c}{MaxFPS}   & \multicolumn{1}{c}{} & \multicolumn{2}{c}{SATLike3.0} \\ \cline{3-4} \cline{6-7} 
                                        &                          & \#win.       & time         &                      & \#win.           & time        \\ \hline
aes                                     & 6                        & \textbf{5}   & 37.65           &                      & 2                & 136.81      \\
atcoss                                  & 14                       & \textbf{1}   & 272.39          &                      & 0                & 0.00        \\
decision-tree                           & 23                       & \textbf{21}  & 17.76           &                      & 6                & 115.36      \\
extension-enforcement                   & 19                       & 14           & 96.19           &                      & \textbf{17}      & 100.30      \\
gen-hyper-tw                            & 37                       & 26           & \textbf{103.40} &                      & 26               & 110.05      \\
hs-timetabling                          & 1                        & \textbf{1}   & 1.59            &                      & 0                & 0.00        \\
large-graph-commmunity                  & 3                        & \textbf{3}   & 6.76            &                      & 2                & 10.51       \\
logic-synthesis                         & 1                        & \textbf{1}   & 2.57            &                      & 0                & 0.00        \\
bcp                                     & 24                       & \textbf{22}  & 65.85           &                      & 6                & 129.55      \\
pseudoBoolean                           & 11                       & \textbf{1}   & 296.65          &                      & 0                & 0.00        \\
maxclique \& maxcut                     & 58                       & \textbf{58}  & 12.87           &                      & 54               & 1.42        \\
MCS-GE        & 25                       & \textbf{25}  & 9.24            &                      & 14               & 32.56       \\
MaxSATQIC & 35                       & \textbf{25}  & 40.29           &                      & 19               & 55.75       \\
min-fill                                & 16                       & \textbf{11}  & 28.78           &                      & 6                & 84.63       \\
optic                                   & 17                       & \textbf{17}  & 39.89           &                      & 0                & 0.00        \\
phylogenetic-trees                      & 11                       & \textbf{2}   & 131.40          &                      & 0                & 0.00        \\
railroad\_reisch                        & 9                        & \textbf{9}   & 39.14           &                      & 5                & 6.57        \\
railway-transport                       & 4                        & \textbf{2}   & 34.92           &                      & 1                & 298.83      \\
ramsey                                  & 14                       & 14           & \textbf{0.04}   &                      & 14               & 0.11        \\
reversi                                 & 11                       & \textbf{2}   & 104.29          &                      & 0                & 0.00        \\
des                                     & 13                       & 1            & 225.94          &                      & \textbf{2}       & 89.65       \\
scheduling                              & 5                        & 2            & 84.98           &                      & \textbf{3}       & 154.90      \\
scheduling\_xiaojuan                    & 8                        & \textbf{5}   & 127.53          &                      & 4                & 85.77       \\
set-covering                            & 9                        & \textbf{9}   & 68.72           &                      & 1                & 122.93      \\
setcover-rail\_zhendong                 & 4                        & \textbf{4}   & 210.15          &                      & 2                & 1.72        \\
treewidth-computation                   & 9                        & \textbf{9}   & 134.01          &                      & 5                & 65.29       \\
uaq                                     & 20                       & \textbf{20}  & 11.10           &                      & 18               & 46.98       \\
uaq\_gazzarata                          & 1                        & \textbf{1}   & 188.66          &                      & 0                & 0.00        \\
xai-mindset2                            & 19                       & \textbf{16}  & 44.07           &                      & 1                & 227.51      \\
mbd                                     & 6                        & 3            & 141.53          &                      & \textbf{5}       & 194.20      \\
SeanSafarpour                           & 13                       & \textbf{9}   & 141.25          &                      & 8                & 142.53      \\
fault-diagnosis                         & 8                        & \textbf{7}   & 31.42           &                      & 0                & 0.00        \\
close\_solutions                        & 14                       & 5            & 43.89           &                      & \textbf{9}       & 35.75       \\
causal-discovery                        & 3                        & 3            & \textbf{3.27}   &                      & 3                & 5.40        \\ \hline
Total                                   & 471                      & \textbf{354} & 48.93           &                      & 233              & 59.02      \\ \toprule
\end{tabular}}
\caption{Comparison of MaxFPS and SATLike3.0 on each PMS instance class. MCS-GE (resp. MaxSATQIC) is a short name of MaximumCommonSub-GraphExtraction (resp. MaxSATQueriesinInterpretableClassifiers).}
\label{table_PMS_class}
\end{table}
\begin{table}[t]
\footnotesize
\centering
\scalebox{0.83}{\begin{tabular}{lrrrrrr} \bottomrule
\multirow{2}{*}{Benchmark}         & \multirow{2}{*}{\#inst.} & \multicolumn{2}{c}{MaxFPS}    &  & \multicolumn{2}{c}{SATLike3.0} \\ \cline{3-4} \cline{6-7} 
                                        &                          & \#win.       & time           &  & \#win.           & time        \\ \hline
abstraction-refinement                  & 10                       & \textbf{9}   & 175.95         &  & 2                & 170.69      \\
af-synthesis                            & 32                       & \textbf{28}  & 21.97          &  & 5                & 196.89      \\
correlation-clustering                  & 44                       & 16           & 145.31         &  & \textbf{31}      & 131.22      \\
decision-tree                           & 24                       & 11           & 160.08         &  & \textbf{17}      & 127.94      \\
hs-timetabling                          & 13                       & \textbf{8}   & 146.10         &  & 0                & 0.00        \\
lisbon-wedding                          & 21                       & \textbf{15}  & 132.27         &  & 0                & 0.00        \\
maxcut                                  & 28                       & \textbf{28}  & 6.20           &  & 26               & 1.68        \\
MaxSATQIC & 32                       & \textbf{26}  & 81.67          &  & 9                & 48.93       \\
metro                                   & 2                        & \textbf{2}   & 234.82         &  & 0                & 0.00        \\
MWDSP       & 7                        & 6            & \textbf{90.34} &  & 6                & 105.84      \\
min-width                               & 40                       & \textbf{39}  & 146.55         &  & 1                & 153.18      \\
mpe                                     & 19                       & \textbf{19}  & 74.17          &  & 5                & 91.75       \\
RBAC                                    & 54                       & 23           & 140.27         &  & \textbf{33}      & 134.35      \\
railroad\_reisch                        & 6                        & \textbf{6}   & 103.38         &  & 1                & 17.14       \\
railway-transport                       & 4                        & 1            & 46.68          &  & \textbf{2}       & 107.02      \\
ramsey                                  & 12                       & 9            & 31.09          &  & \textbf{11}      & 31.70       \\
relational-inference                    & 2                        & 1            & \textbf{73.91} &  & 1                & 212.09      \\
scSequencing\_Mehrabadi                 & 10                       & 2            & 85.73          &  & \textbf{8}       & 57.30       \\
set-covering                            & 13                       & \textbf{13}  & 68.63          &  & 1                & 42.34       \\
staff-scheduling                        & 11                       & \textbf{11}  & 154.20         &  & 0                & 0.00        \\
spot5                                   & 5                        & \textbf{5}   & 82.66          &  & 0                & 0.00        \\
causal-discovery                        & 24                       & \textbf{23}  & 19.15          &  & 15               & 17.64       \\
timetabling                             & 19                       & \textbf{15}  & 187.30         &  & 0                & 0.00        \\
max-realizability                       & 13                       & \textbf{10}  & 18.43          &  & 8                & 65.04       \\
BTBNSL-Rounded                          & 26                       & \textbf{14}  & 22.85          &  & 12               & 64.80       \\
tcp                                     & 13                       & \textbf{12}  & 128.59         &  & 2                & 241.91      \\
cluster-expansion                       & 20                       & 7            & 0.04           &  & \textbf{17}      & 81.70       \\ \hline
Total                                   & 504                      & \textbf{359} & 90.77          &  & 213              & 86.67      \\ \toprule
\end{tabular}}
\caption{Comparison of MaxFPS and SATLike3.0 on each WPMS instance class. MWDSP (resp. MaxSATQIC) is a short name of MinimumWeightDominatingSetProblem (resp. MaxSATQueriesinInterpretableClassifiers).}
\label{table_WPMS_class}
\end{table}
\begin{table}[t]
\footnotesize
\centering
\begin{tabular}{lrrrrrr} \bottomrule
\multicolumn{1}{l}{\multirow{2}{*}{Benchmark}} & \multirow{2}{*}{\#inst.} & \multicolumn{2}{c}{CCEHC-FPS} &  & \multicolumn{2}{c}{CCEHC} \\ \cline{3-4} \cline{6-7}
\multicolumn{1}{c}{}                           &                          & \#win.           & time       &  & \#win.         & time     \\ \hline
WPMS\_2018                                     & 172                      & \textbf{78}      & 101.40     &  & 58             & 85.49    \\
WPMS\_2019                                     & 297                      & \textbf{143}     & 114.06     &  & 94             & 98.32    \\
WPMS\_2020                                     & 253                      & \textbf{112}     & 129.33     &  & 89             & 129.43   \\
WPMS\_2021                                     & 151                      & 46               & 157.44     &  & \textbf{59}    & 136.29  \\ \toprule
\end{tabular}
\caption{Comparison of CCEHC-FPS and CCEHC.}
\label{table_CCEHC}
\end{table}
\begin{table}[!t]
\footnotesize
\centering
\begin{tabular}{lrrrrrr} \bottomrule
\multicolumn{1}{l}{\multirow{2}{*}{Benchmark}} & \multirow{2}{*}{\#inst.} & \multicolumn{2}{c}{Dist-FPS} &  & \multicolumn{2}{c}{Dist} \\ \cline{3-4} \cline{6-7}
\multicolumn{1}{c}{}                           &                          & \#win.           & time      &  & \#win.      & time       \\ \hline
PMS\_2018                                      & 153                      & \textbf{87}      & 92.43     &  & 73          & 93.47      \\
PMS\_2019                                      & 299                      & \textbf{170}     & 62.73     &  & 152         & 77.47      \\
PMS\_2020                                      & 262                      & \textbf{145}     & 62.16     &  & 119         & 84.77      \\
PMS\_2021                                      & 155                      & \textbf{77}      & 87.17     &  & 74          & 71.68       \\ \toprule
\end{tabular}
\caption{Comparison of Dist-FPS and Dist.}
\label{table_Dist}
\end{table}
\begin{table}[t]
\footnotesize
\centering
\begin{tabular}{lrrrrrr} \bottomrule
\multirow{2}{*}{Benchmark} & \multirow{2}{*}{\hspace{-0.5em}\#inst.} & \multicolumn{2}{c}{BandMS-FPS} &  & \multicolumn{2}{c}{BandMaxSAT} \\ \cline{3-4} \cline{6-7}
                           &                          & \#win.              & time         &  & \#win.         & time          \\ \hline
PMS\_2018                  & \hspace{-0.5em}153                      & \textbf{105}        & 70.91        &  & 79             & 102.39        \\
PMS\_2019                  & \hspace{-0.5em}299                      & \textbf{203}        & 68.81        &  & 158            & 67.44         \\
PMS\_2020                  & \hspace{-0.5em}262                      & \textbf{172}        & 65.39        &  & 135            & 80.15         \\
PMS\_2021                  & \hspace{-0.5em}155                      & \textbf{91}         & 74.66        &  & 89             & 67.78         \\
WPMS\_2018                 & \hspace{-0.5em}172                      & \textbf{102}        & 105.83       &  & 73             & 104.27        \\
WPMS\_2019                 & \hspace{-0.5em}297                      & \textbf{179}        & 98.49       &  & 149            & 89.74         \\
WPMS\_2020                 & \hspace{-0.5em}253                      & \textbf{153}        & 111.28       &  & 121            & 122.00        \\
WPMS\_2021                 & \hspace{-0.5em}151                      & \textbf{83}         & 132.67       &  & 62             & 133.11   \\ \toprule    
\end{tabular}
\caption{Comparison of BandMS-FPS and BandMaxSAT.}
\label{table_BandMaxSAT}
\end{table}
\begin{table}[!t]
\footnotesize
\centering
\begin{tabular}{lrrrr} \bottomrule
Benchmark  & \hspace{-1em}MaxFPS-c        & SATLike-c & TT-OWI & Loandra \\ \hline
PMS\_2018  & \hspace{-1em}\textbf{0.8682} & 0.8399    & 0.8371 & 0.7912  \\
PMS\_2019  & \hspace{-1em}\textbf{0.8796} & 0.8734    & 0.8698 & 0.7919  \\
PMS\_2020  & \hspace{-1em}\textbf{0.8696} & 0.8457    & 0.8534 & 0.8156  \\
PMS\_2021  & \hspace{-1em}\textbf{0.8802} & 0.8508    & 0.8407 & 0.8127  \\
WPMS\_2018 & \hspace{-1em}\textbf{0.9084} & 0.8931    & 0.8994 & 0.8752  \\
WPMS\_2019 & \hspace{-1em}\textbf{0.9091} & 0.8810    & 0.9010 & 0.8321  \\
WPMS\_2020 & \hspace{-1em}\textbf{0.8868} & 0.8609    & 0.8650 & 0.8312  \\
WPMS\_2021 & \hspace{-1em}\textbf{0.8101} & 0.7843    & 0.7758 & 0.7945 \\ \toprule
\end{tabular}
\caption{Comparison of MaxFPS-c and complete solvers.}
\label{table_MaxFPSc}
\end{table}
\begin{table}[t]
\footnotesize
\centering
\begin{tabular}{lrrrrrr} \bottomrule
\multirow{2}{*}{Benchmark} & \multirow{2}{*}{\#inst.} & \multicolumn{2}{c}{MaxFPS} &  & \multicolumn{2}{c}{MaxFPS$_1$} \\ \cline{3-4} \cline{6-7} 
                           &                          & \#win.          & time     &  & \#win.         & time         \\ \hline
PMS\_2018                  & 153                      & \textbf{104}    & 71.21    &  & 75             & 82.91        \\
PMS\_2019                  & 299                      & \textbf{202}    & 55.93    &  & 172            & 55.63        \\
PMS\_2020                  & 262                      & \textbf{163}    & 48.37    &  & 140            & 67.44        \\
PMS\_2021                  & 155                      & \textbf{98}     & 53.10    &  & 89             & 46.67        \\
WPMS\_2018                 & 172                      & \textbf{119}    & 69.45    &  & 69             & 49.45        \\
WPMS\_2019                 & 297                      & \textbf{212}    & 92.88    &  & 123            & 77.06        \\
WPMS\_2020                 & 253                      & \textbf{177}    & 90.82    &  & 104            & 83.44        \\
WPMS\_2021                 & 151                      & \textbf{89}     & 107.00   &  & 61             & 83.44       \\ \toprule
\end{tabular}
\caption{Comparison of MaxFPS and MaxFPS$_1$.}
\label{table_V1}
\end{table}
\begin{table}[t]
\footnotesize
\centering
\begin{tabular}{lrrrrrr} \bottomrule
\multirow{2}{*}{Benchmark} & \multirow{2}{*}{\#inst.} & \multicolumn{2}{c}{MaxFPS$_1$} &  & \multicolumn{2}{c}{SATLike3.0} \\ \cline{3-4} \cline{6-7} 
                           &                          &  \#win.            & time      &  & \#win.           & time        \\ \hline
PMS\_2018                  & 153                      &  \textbf{106}      & 72.42     &  & 63               & 83.74       \\
PMS\_2019                  & 299                      &  \textbf{206}      & 51.51     &  & 142              & 53.88       \\
PMS\_2020                  & 262                      &  \textbf{161}      & 53.66     &  & 112              & 75.61       \\
PMS\_2021                  & 155                      &  \textbf{105}      & 46.68     &  & 68               & 62.79       \\
WPMS\_2018                 & 172                      &  90                & 74.04     &  & \textbf{92}      & 90.23       \\
WPMS\_2019                 & 297                      &  \textbf{188}      & 88.56     &  & 131              & 96.44       \\
WPMS\_2020                 & 253                      &  \textbf{156}      & 90.30     &  & 116              & 98.30       \\
WPMS\_2021                 & 151                      &  71                & 86.66     &  & \textbf{75}      & 109.47     \\ \toprule
\end{tabular}
\caption{Comparison of MaxFPS$_1$ and SATLike3.0.}
\label{table_V1_2}
\end{table}
\begin{table}[!t]
\footnotesize
\centering
\begin{tabular}{lrrrrrrrrr} \bottomrule
\multirow{2}{*}{Benchmark} &  \multirow{2}{*}{\hspace{-1em}\#inst.} &  \multicolumn{2}{c}{\hspace{-1em}MaxFPS} & \hspace{-1em}  &  \multicolumn{2}{c}{\hspace{-1em}MaxFPS$_2$} & \hspace{-1em}  &  \multicolumn{2}{c}{\hspace{-1em}MaxFPS$_3$} \\ \cline{3-4} \cline{6-7} \cline{9-10} 
                           & \hspace{-1em}                          & \hspace{-1.2em} \#win.          & \hspace{-1em} time     & \hspace{-1em}  & \hspace{-1.2em} \#win.        & \hspace{-1em} time        & \hspace{-1em}  & \hspace{-1.2em} \#win.       & \hspace{-1em} time         \\ \hline
PMS\_2018                  & \hspace{-1.2em} 153                      & \hspace{-1.2em} \textbf{96}     & \hspace{-1.2em} 78.45    & \hspace{-1.2em}  & \hspace{-1.2em} 44            & \hspace{-1.2em} 20.69       & \hspace{-1.2em}  & \hspace{-1.2em} 80           & \hspace{-1.2em} 59.52        \\
PMS\_2019                  & \hspace{-1.2em} 299                      & \hspace{-1.2em} \textbf{194}    & \hspace{-1.2em} 62.70    & \hspace{-1.2em}  & \hspace{-1.2em} 103           & \hspace{-1.2em} 13.05       & \hspace{-1.2em}  & \hspace{-1.2em} 166          & \hspace{-1.2em} 46.30        \\
PMS\_2020                  & \hspace{-1.2em} 262                      & \hspace{-1.2em} \textbf{148}    & \hspace{-1.2em} 50.68    & \hspace{-1.2em}  & \hspace{-1.2em} 86            & \hspace{-1.2em} 16.96       & \hspace{-1.2em}  & \hspace{-1.2em} 136          & \hspace{-1.2em} 63.19        \\
PMS\_2021                  & \hspace{-1.2em} 155                      & \hspace{-1.2em} \textbf{89}     & \hspace{-1.2em} 59.75    & \hspace{-1.2em}  & \hspace{-1.2em} 58            & \hspace{-1.2em} 13.01       & \hspace{-1.2em}  & \hspace{-1.2em} 85           & \hspace{-1.2em} 39.88        \\
WPMS\_2018                 & \hspace{-1.2em} 172                      & \hspace{-1.2em} \textbf{85}     & \hspace{-1.2em} 66.54    & \hspace{-1.2em}  & \hspace{-1.2em} 57            & \hspace{-1.2em} 13.23       & \hspace{-1.2em}  & \hspace{-1.2em} 70           & \hspace{-1.2em} 74.63        \\
WPMS\_2019                 & \hspace{-1.2em} 297                      & \hspace{-1.2em} \textbf{181}    & \hspace{-1.2em} 91.87    & \hspace{-1.2em}  & \hspace{-1.2em} 92            & \hspace{-1.2em} 13.28       & \hspace{-1.2em}  & \hspace{-1.2em} 144          & \hspace{-1.2em} 78.45        \\
WPMS\_2020                 & \hspace{-1.2em} 253                      & \hspace{-1.2em} \textbf{164}    & \hspace{-1.2em} 94.40    & \hspace{-1.2em}  & \hspace{-1.2em} 61            & \hspace{-1.2em} 12.52       & \hspace{-1.2em}  & \hspace{-1.2em} 112          & \hspace{-1.2em} 75.66        \\
WPMS\_2021                 & \hspace{-1.2em} 151                      & \hspace{-1.2em} \textbf{83}     & \hspace{-1.2em} 118.80   & \hspace{-1.2em}  & \hspace{-1.2em} 32            & \hspace{-1.2em} 21.86       & \hspace{-1.2em}  & \hspace{-1.2em} 51           & \hspace{-1.2em} 103.99 \\ \toprule     
\end{tabular}
\caption{Comparison of MaxFPS, MaxFPS$_2$, MaxFPS$_3$.}
\label{table_V2}
\end{table}

The comparison results of MaxFPS and SATLike3.0 are shown in Table \ref{table_MaxFPS}. Column \textit{\#inst.} indicates the number of instances of each benchmark, column \textit{\#win.} indicates the number of instances in which the solver finds the best solution among all solvers in the table, and column \textit{time} indicates the average time (in seconds) for obtaining the results of the \textit{winning} instances. We could observe that the \textit{winning} PMS (resp. WPMS) instances of MaxFPS are about 49-82\% (resp. 31-122\%) greater than those of SATLike3.0, indicating a significant improvement.

To obtain a more detailed comparison between MaxFPS and SATLike3.0 and evaluate the performance of FPS on different instance classes, we collect all the tested instances (duplicated ones are removed) and compare MaxFPS with SATLike3.0 on each instance class. Ties of these two algorithms with the same number of \textit{winning} instances are broken by selecting the one with less running time (as the rules in MSEs). The results on the PMS and WPMS instance classes are shown in Tables \ref{table_PMS_class} and \ref{table_WPMS_class}, respectively. Note that we remove the instance classes that both MaxFPS and SATLike3.0 can not yield feasible solutions. The results show that MaxFPS outperforms SATLike3.0 on most classes of both PMS and WPMS instances. Specifically, for all the 34 (resp. 27) classes of PMS (resp. WPMS) instances, MaxFPS outperforms SATLike3.0 on 29 (resp. 20) classes, indicating the excellent robustness of FPS that can boost SATLike3.0 in solving various classes of (W)PMS instances. 

The comparison results of CCEHC-FPS and CCEHC, Dist-FPS and Dist, BandMaxSAT-FPS (BandMS-FPS) and BandMaxSAT, are summarized in Tables \ref{table_CCEHC}, \ref{table_Dist}, and \ref{table_BandMaxSAT}, respectively. The results show that these baselines can all be significantly improved by our FPS strategy, indicating its excellent generalization performance and robustness. Moreover, FPS can also significantly improve the SATLike3.0, CCEHC, Dist, and BandMaxSAT algorithms within 60 seconds of time limit (see details in Appendix).

\subsection{Comparison with SAT-based (W)PMS Solvers}
We then replace the local search component in SATLike-c (\ie, SATLike3.0) with MaxFPS, and compare the resulting solver MaxFPS-c with some of the state-of-the-art SAT-based (W)PMS solvers, including SATLike-c, TT-Open-WBO-Inc (TT-OWI), and Loandra. Note that the results of these SAT-based solvers are obtained by running their codes downloaded from MSE2021. We use the scoring function in MSEs (introduced in Section \ref{sec-para}) to calculate a score for each solver per instance, and present the average score of each solver on each benchmark in Table \ref{table_MaxFPSc}. We also depict the distribution of scores per instance of these SAT-based solvers as in MSEs in Appendix. The results show that MaxFPS-c yields the highest average score on all the benchmarks. Moreover, FPS not only can improve SATLike-c, but also help it outperform the other SAT-based (W)PMS solvers. These results further indicate the excellent performance of FPS.

\subsection{Ablation Study}
Finally, we do ablation studies to evaluate the effectiveness and rationality of components in FPS, including the local optima escaping method and the two-level look-ahead search process, by comparing MaxFPS with its several variants.

To evaluate the effectiveness of the components, we first perform two groups of comparison. The first one compares MaxFPS with its variant, MaxFPS$_1$, which replaces the local optima escaping strategy in MaxFPS (lines 28-29 in Algorithm \ref{alg_FPS}) with the simple random walk strategy (lines 8-9 in Algorithm \ref{alg_single}). The second one compares MaxFPS$_1$ with SATLike3.0. The results of these two groups of comparison are shown in Tables \ref{table_V1} and \ref{table_V1_2}, respectively. We can observe that MaxFPS significantly outperforms MaxFPS$_1$, indicating that the local optima escaping strategy in FPS is significantly better than the random walk strategy that is widely used in recent local search (W)PMS algorithms. This is because our method can provide more and better directions to escape from local optima. Moreover, MaxFPS$_1$ significantly outperforms SATLike3.0, indicating that only using the two-level look-ahead method can also significantly improve the local search algorithms, by improving the local optimal solutions of the single flipping mechanism. 


Then, we compare MaxFPS with its other two variants, MaxFPS$_2$ and MaxFPS$_3$, to analyze the rationality of the two-level search process in MaxFPS. MaxFPS$_2$ extends the two-level search approach in MaxFPS to three levels. In MaxFPS$_2$, the third-level variables are selected by the same approach for selecting the second-level variables as in MaxFPS, and the second-level variables in MaxFPS$_2$ are sampled from the neighbors (variables that appeared in the same clause) of the corresponding first-level variable. MaxFPS$_3$ is a variant of MaxFPS that also look-ahead when the current solution is not a local optimum for the single flipping mechanism, \ie, $GoodVars \neq \emptyset$. The results of MaxFPS and these two variants are summarised in Table \ref{table_V2}.

From the results in Table \ref{table_V2}, we can see that MaxFPS significantly outperforms MaxFPS$_2$. This is because the three-level look-ahead process is too time-consuming and inefficient. Thus the proposed two-level look-ahead technique in FPS is reasonable. We can also observe that MaxFPS outperforms MaxFPS$_3$, demonstrating that look-ahead only when $GoodVars = \emptyset$ is reasonable and effective.

\section{Conclusion}
In this work, we propose an effective and general strategy to replace the mechanism of flipping a single variable per iteration that is widely used in local search MaxSAT solvers. The proposed FPS strategy combines the look-ahead technique and probabilistic sampling method. As a result, FPS can improve the local optimum of the single flipping mechanism and provide more and better search directions to escape from local optima.


Look-ahead is not a new technique, but how to look-ahead is the magic recipe. This work proposes an effective way to apply this technique to boost local search MaxSAT solvers. We also demonstrate that there is great potential for the look-ahead technique to be used for MaxSAT. Extensive experiments demonstrate that our proposed FPS strategy significantly improves the state-of-the-art (W)PMS solvers, and FPS has an excellent generalization performance to various local search MaxSAT solvers, including SATLike3.0, CCEHC, Dist, and BandMaxSAT as well as one of the state-of-the-art SAT-based solvers, SATLike-c.

Moreover, we have also tried to apply FPS to local search SAT solvers such as CCAnr~\cite{Cai2015-CCAnr} and obtained promising performance. In the future, we will further explore the potential of FPS in MaxSAT and SAT solving.

\section*{Acknowledgments}
This work is supported by National Natural Science Foundation (U22B2017) and MSRA Collaborative Research 2022 (100338928).

\bibliography{aaai23}

\appendix

\newpage
\twocolumn[
\begin{@twocolumnfalse}
\section*{\centering{\LARGE{Appendix}}}
~\\
\end{@twocolumnfalse}
]

In the Appendix, we mainly present some supplementary experiments. We first present the comprehensive comparison results between the algorithms improved by FPS and the local search (W)PMS baselines within two time limits, 60 and 300 seconds, as in MSEs. Then, we present the detailed comparison of MaxFPS-c and the SAT-based solvers. Finally, we present some additional ablation studies.

Benchmarks used in Appendix are the same as those described in Section 4.1, \ie, all the (W)PMS instances from the incomplete track of the last four MSEs.

\begin{table}[!b]
\footnotesize
\centering
\begin{tabular}{lrrrrrr} \bottomrule
\multicolumn{1}{l}{\multirow{2}{*}{Benchmark}} & \multirow{2}{*}{\#inst.} & \multicolumn{2}{c}{MaxFPS} &  & \multicolumn{2}{c}{SATLike3.0} \\ \cline{3-4} \cline{6-7} 
\multicolumn{1}{c}{}                           &                          & \#win.          & time     &  & \#win.         & time          \\ \hline
\multicolumn{7}{l}{60 seconds of time limit}                                                                                               \\ \hline
PMS\_2018                                      & 153                      & \textbf{104}    & 13.69    &  & 49             & 15.50         \\
PMS\_2019                                      & 299                      & \textbf{199}    & 11.52    &  & 132            & 11.87         \\
PMS\_2020                                      & 262                      & \textbf{180}    & 12.03    &  & 100            & 10.88         \\
PMS\_2021                                      & 155                      & \textbf{103}    & 9.56     &  & 57             & 8.63          \\
WPMS\_2018                                     & 172                      & \textbf{118}    & 16.90    &  & 62             & 11.97         \\
WPMS\_2019                                     & 297                      & \textbf{209}    & 20.15    &  & 94             & 17.19         \\
WPMS\_2020                                     & 253                      & \textbf{167}    & 19.57    &  & 84             & 20.85         \\
WPMS\_2021                                     & 151                      & \textbf{73}     & 25.75    &  & 58             & 27.66         \\ \hline
\multicolumn{7}{l}{300 seconds of time limit}                                                                                              \\ \hline
PMS\_2018                                      & 153                      & \textbf{111}    & 60.68    &  & 61             & 83.43         \\
PMS\_2019                                      & 299                      & \textbf{212}    & 49.05    &  & 142            & 56.64         \\
PMS\_2020                                      & 262                      & \textbf{189}    & 41.09    &  & 112            & 62.14         \\
PMS\_2021                                      & 155                      & \textbf{108}    & 47.12    &  & 64             & 51.00         \\
WPMS\_2018                                     & 172                      & \textbf{115}    & 78.90    &  & 64             & 84.37         \\
WPMS\_2019                                     & 297                      & \textbf{222}    & 94.71    &  & 100            & 80.38         \\
WPMS\_2020                                     & 253                      & \textbf{183}    & 92.49    &  & 86             & 72.82         \\
WPMS\_2021                                     & 151                      & \textbf{84}     & 104.92   &  & 64             & 98.31      \\ \toprule  
\end{tabular}
\caption{Comparison of MaxFPS and SATLike3.0.}
\label{table_MaxFPS_all}
\end{table}
\begin{table}[!b]
\footnotesize
\centering
\scalebox{0.98}{\begin{tabular}{lrrrrrr} \bottomrule
\multirow{2}{*}{Benchmark} & \multirow{2}{*}{\#inst.} & \multicolumn{2}{c}{CCEHC-FPS} &  & \multicolumn{2}{c}{CCEHC} \\ \cline{3-4} \cline{6-7} 
                           &                          & \#win.           & time       &  & \#win.         & time     \\ \hline
\multicolumn{7}{l}{60 seconds of time limit}                                                                         \\ \hline
WPMS\_2018                 & 172                      & \textbf{77}      & 25.99      &  & 57             & 21.11    \\
WPMS\_2019                 & 297                      & \textbf{138}     & 27.34      &  & 82             & 21.94    \\
WPMS\_2020                 & 253                      & \textbf{96}      & 25.77      &  & 79             & 26.82    \\
WPMS\_2021                 & 151                      & 40               & 32.42      &  & \textbf{48}    & 34.30    \\ \hline
\multicolumn{7}{l}{300 seconds of time limit}                                                                        \\ \hline
WPMS\_2018                 & 172                      & \textbf{78}      & 101.40     &  & 58             & 85.49    \\
WPMS\_2019                 & 297                      & \textbf{143}     & 114.06     &  & 94             & 98.32    \\
WPMS\_2020                 & 253                      & \textbf{112}     & 129.33     &  & 89             & 129.43   \\
WPMS\_2021                 & 151                      & 46               & 157.44     &  & \textbf{59}    & 136.29  \\ \toprule
\end{tabular}}
\caption{Comparison of CCEHC-FPS and CCEHC.}
\label{table_CCEHC_all}
\end{table}
\begin{table}[!b]
\footnotesize
\centering
\begin{tabular}{lrrrrrr} \bottomrule
\multirow{2}{*}{Benchmark} & \multirow{2}{*}{\#inst.} & \multicolumn{2}{c}{Dist-FPS} &  & \multicolumn{2}{c}{Dist} \\ \cline{3-4} \cline{6-7} 
                           &                          & \#win.           & time      &  & \#win.      & time       \\ \hline
\multicolumn{7}{l}{60 seconds of time limit}                                                                       \\ \hline
PMS\_2018                  & 153                      & \textbf{87}      & 20.70     &  & 67          & 19.25      \\
PMS\_2019                  & 299                      & \textbf{168}     & 14.16     &  & 146         & 17.23      \\
PMS\_2020                  & 262                      & \textbf{144}     & 14.50     &  & 110         & 17.52      \\
PMS\_2021                  & 155                      & \textbf{76}      & 18.19     &  & 69          & 11.44      \\ \hline
\multicolumn{7}{l}{300 seconds of time limit}                                                                      \\ \hline
PMS\_2018                  & 153                      & \textbf{87}      & 92.43     &  & 73          & 93.47      \\
PMS\_2019                  & 299                      & \textbf{170}     & 62.73     &  & 152         & 77.47      \\
PMS\_2020                  & 262                      & \textbf{145}     & 62.16     &  & 119         & 84.77      \\
PMS\_2021                  & 155                      & \textbf{77}      & 87.17     &  & 74          & 71.68     \\ \toprule
\end{tabular}
\caption{Comparison of Dist-FPS and Dist.}
\label{table_Dist_all}
\end{table}
\begin{table}[!b]
\footnotesize
\centering
\scalebox{0.96}{\begin{tabular}{lrrrrrr} \bottomrule
\multirow{2}{*}{Benchmark} & \multirow{2}{*}{\#inst.} & \multicolumn{2}{c}{BandMS-FPS} &  & \multicolumn{2}{c}{BandMaxSAT} \\ \cline{3-4} \cline{6-7} 
                           &                          & \#win.            & time       &  & \#win.           & time        \\ \hline
\multicolumn{7}{l}{60 seconds of time limit}                                                                               \\ \hline
PMS\_2018                  & 153                      & \textbf{91}       & 15.70      &  & 76               & 19.62       \\
PMS\_2019                  & 299                      & \textbf{182}      & 14.36      &  & 157              & 15.43       \\
PMS\_2020                  & 262                      & \textbf{160}      & 13.47      &  & 134              & 17.76       \\
PMS\_2021                  & 155                      & 79                & 12.69      &  & \textbf{91}      & 17.61       \\
WPMS\_2018                 & 172                      & \textbf{102}      & 20.43      &  & 82               & 20.44       \\
WPMS\_2019                 & 297                      & \textbf{165}      & 20.15      &  & 151              & 23.16       \\
WPMS\_2020                 & 253                      & \textbf{137}      & 21.96      &  & 122              & 23.66       \\
WPMS\_2021                 & 151                      & \textbf{72}       & 29.78      &  & 63               & 30.57       \\ \hline
\multicolumn{7}{l}{300 seconds of time limit}                                                                              \\ \hline
PMS\_2018                  & 153                      & \textbf{105}      & 70.91        &  & 79             & 102.39        \\
PMS\_2019                  & \hspace{-0.5em}299                      & \textbf{203}        & 68.81        &  & 158            & 67.44         \\
PMS\_2020                  & \hspace{-0.5em}262                      & \textbf{172}        & 65.39        &  & 135            & 80.15         \\
PMS\_2021                  & \hspace{-0.5em}155                      & \textbf{91}         & 74.66        &  & 89             & 67.78         \\
WPMS\_2018                 & \hspace{-0.5em}172                      & \textbf{102}        & 105.83       &  & 73             & 104.27        \\
WPMS\_2019                 & \hspace{-0.5em}297                      & \textbf{179}        & 98.49       &  & 149            & 89.74         \\
WPMS\_2020                 & \hspace{-0.5em}253                      & \textbf{153}        & 111.28       &  & 121            & 122.00        \\
WPMS\_2021                 & \hspace{-0.5em}151                      & \textbf{83}         & 132.67       &  & 62             & 133.11     \\ \toprule
\end{tabular}}
\caption{Comparison of BandMS-FPS and BandMaxSAT.}
\label{table_BandMaxSAT_all}
\end{table}

\section{Comprehensive Evaluation on FPS}
The comparison results, within 60 or 300 seconds of time limit, of MaxFPS and SATLike3.0~\cite{Cai2020}, CCEHC-FPS and CCEHC~\cite{Luo2017}, Dist-FPS and Dist~\cite{Cai2014}, BandMaxSAT-FPS (BandMS-FPS) and BandMaxSAT~\cite{Zheng2022}, are summarized in Tables \ref{table_MaxFPS_all}, \ref{table_CCEHC_all}, \ref{table_Dist_all}, and \ref{table_BandMaxSAT_all}, respectively. The results show that, within either 60 or 300 seconds of time limit, FPS can significantly improve these state-of-the-art local search (W)PMS algorithms, indicating its excellent performance and robustness.

\begin{table}[!t]
\footnotesize
\centering
\scalebox{0.83}{\begin{tabular}{lrrrrrr} \bottomrule
\multirow{2}{*}{Benchmark}         & \multirow{2}{*}{\#inst.} & \multicolumn{2}{c}{MaxFPS}    &           & \multicolumn{2}{c}{SATLike3.0} \\ \cline{3-4} \cline{6-7} 
                                        &                          & \#win.       & time           &           & \#win.       & time            \\ \hline
aes                                     & 6                        & \textbf{4}   & 8.40           &           & 1            & 18.91           \\
decision-tree                           & 23                       & \textbf{20}  & 7.54           &           & 3            & 10.27           \\
extension-enforcement                   & 19                       & 14           & \textbf{14.86} & \textbf{} & 14           & 19.44           \\
gen-hyper-tw                            & 37                       & 29           & 19.28          &           & \textbf{20}  & 15.90           \\
hs-timetabling                          & 1                        & \textbf{1}   & 1.59           &           & 0            & 0.00            \\
large-graph-commmunity                  & 3                        & \textbf{3}   & 6.76           &           & 2            & 10.51           \\
logic-synthesis                         & 1                        & \textbf{1}   & 2.57           &           & 0            & 0.00            \\
bcp                                     & 24                       & \textbf{19}  & 21.06          &           & 5            & 35.71           \\
pseudoBoolean                           & 11                       & \textbf{1}   & 2.87           &           & 0            & 0.00            \\
maxclique \& maxcut                     & 58                       & 55           & 3.44           &           & 55           & \textbf{1.40}   \\
MCS-GE        & 25                       & \textbf{25}  & 7.97           &           & 11           & 3.53            \\
MaxSATQIC & 35                       & \textbf{27}  & 12.55          &           & 18           & 8.98            \\
min-fill                                & 16                       & \textbf{10}  & 15.82          &           & 6            & 32.44           \\
optic                                   & 17                       & \textbf{15}  & 11.18          &           & 2            & 12.73           \\
phylogenetic-trees                      & 11                       & \textbf{1}   & 36.80          &           & 0            & 0.00            \\
railroad\_reisch                        & 9                        & \textbf{9}   & 9.73           &           & 6            & 5.67            \\
railway-transport                       & 4                        & \textbf{2}   & 34.92          &           & 1            & 37.35           \\
ramsey                                  & 14                       & 14           & \textbf{0.04}  & \textbf{} & 14           & 0.11            \\
des                                     & 13                       & 1            & 59.40          &           & 1            & \textbf{36.59}  \\
scheduling                              & 5                        & \textbf{3}   & 38.29          &           & 2            & 29.83           \\
scheduling\_xiaojuan                    & 8                        & \textbf{6}   & 22.53          &           & 4            & 34.80           \\
set-covering                            & 9                        & \textbf{9}   & 19.75          &           & 0            & 0.00            \\
setcover-rail\_zhendong                 & 4                        & 2            & \textbf{1.53}  & \textbf{} & 2            & 1.72            \\
treewidth-computation                   & 9                        & 5            & 7.94           &           & \textbf{7}   & 20.15           \\
uaq                                     & 20                       & \textbf{20}  & 8.04           &           & 11           & 12.86           \\
uaq\_gazzarata                          & 1                        & \textbf{1}   & 24.91          &           & 0            & 0.00            \\
xai-mindset2                            & 19                       & \textbf{14}  & 6.63           &           & 1            & 30.61           \\
mbd                                     & 6                        & 2            & 23.39          &           & \textbf{4}   & 15.88           \\
SeanSafarpour                           & 13                       & 8            & 15.28          &           & \textbf{10}  & 17.51           \\
fault-diagnosis                         & 8                        & \textbf{7}   & 31.42          &           & 0            & 0.00            \\
close\_solutions                        & 14                       & 5            & 5.61           &           & \textbf{9}   & 18.08           \\
causal-discovery                        & 3                        & 3            & \textbf{3.27}  & \textbf{} & 3            & 5.40            \\ \hline
Total                                   & 446                      & \textbf{336} & 11.49          &           & 212          & 11.22       \\ \toprule   
\end{tabular}}
\caption{Comparison of MaxFPS and SATLike3.0 within 60 seconds of time limit on each PMS instance class. MCS-GE (resp. MaxSATQIC) is a short name of MaximumCommonSub-GraphExtraction (resp. MaxSATQueriesinInterpretableClassifiers).}
\label{table_PMS_class_60s}
\end{table}
\begin{table}[!t]
\footnotesize
\centering
\scalebox{0.83}{\begin{tabular}{lrrrrrr} \bottomrule
\multirow{2}{*}{Benchmark}         & \multirow{2}{*}{\#inst.} & \multicolumn{2}{c}{MaxFPS}    &           & \multicolumn{2}{c}{SATLike3.0} \\ \cline{3-4} \cline{6-7} 
                                        &                          & \#win.       & time           &           & \#win.       & time            \\ \hline
abstraction-refinement                  & 10                       & \textbf{5}   & 44.81          &           & 2            & 58.29           \\
af-synthesis                            & 32                       & \textbf{32}  & 3.70           &           & 1            & 0.21            \\
correlation-clustering                  & 44                       & 21           & 20.18          &           & \textbf{33}  & 22.05           \\
decision-tree                           & 24                       & 4            & 30.97          &           & \textbf{20}  & 38.12           \\
hs-timetabling                          & 13                       & \textbf{7}   & 44.19          &           & 0            & 0.00            \\
lisbon-wedding                          & 21                       & \textbf{15}  & 34.76          &           & 0            & 0.00            \\
maxcut                                  & 28                       & \textbf{27}  & 0.86           &           & 26           & 1.68            \\
MaxSATQIC & 32                       & \textbf{23}  & 18.88          &           & 10           & 18.06           \\
metro                                   & 2                        & 1            & 55.85          &           & 1            & \textbf{48.51}  \\
MWDSP       & 7                        & 2            & \textbf{31.05} & \textbf{} & 2            & 38.79           \\
min-width                               & 40                       & \textbf{39}  & 36.33          &           & 1            & 47.49           \\
mpe                                     & 19                       & \textbf{18}  & 1.86           &           & 3            & 27.52           \\
RBAC                                    & 54                       & 24           & 22.74          &           & \textbf{34}  & 27.64           \\
railroad\_reisch                        & 6                        & \textbf{6}   & 39.91          &           & 1            & 17.14           \\
railway-transport                       & 4                        & \textbf{2}   & 52.84          &           & 1            & 59.01           \\
ramsey                                  & 12                       & 9            & 5.29           &           & \textbf{11}  & 7.79            \\
relational-inference                    & 2                        & \textbf{1}   & 58.97          &           & 0            & 0.00            \\
scSequencing\_Mehrabadi                 & 10                       & 2            & 33.86          &           & \textbf{6}   & 14.39           \\
set-covering                            & 13                       & \textbf{13}  & 12.44          &           & 1            & 42.34           \\
staff-scheduling                        & 11                       & \textbf{10}  & 29.94          &           & 1            & 39.73           \\
spot5                                   & 5                        & \textbf{5}   & 43.68          &           & 0            & 0.00            \\
causal-discovery                        & 24                       & \textbf{23}  & 4.44           &           & 15           & 4.56            \\
timetabling                             & 19                       & \textbf{13}  & 40.21          &           & 1            & 43.44           \\
max-realizability                       & 13                       & \textbf{10}  & 10.26          &           & 4            & 11.64           \\
BTBNSL-Rounded                          & 26                       & \textbf{14}  & 0.40           &           & 12           & 17.06           \\
tcp                                     & 13                       & \textbf{10}  & 22.53          &           & 4            & 15.52           \\
cluster-expansion                       & 20                       & 12           & 0.04           &           & \textbf{14}  & 0.08            \\ \hline
Total                                   & 504                      & \textbf{348} & 18.53          &           & 204          & 18.55    \\ \toprule      
\end{tabular}}
\caption{Comparison of MaxFPS and SATLike3.0 within 60 seconds of time limit on each WPMS instance class. MWDSP (resp. MaxSATQIC) is a short name of MinimumWeightDominatingSetProblem (resp. MaxSATQueriesinInterpretableClassifiers).}
\label{table_WPMS_class_60s}
\end{table}

We also compare MaxFPS and SATLike3.0 within 60 seconds of time limit on each instance class. The results on the PMS and WPMS instance classes are shown in Tables \ref{table_PMS_class_60s} and \ref{table_WPMS_class_60s}, respectively. Note that duplicated instances and the instance classes that both MaxFPS and SATLike3.0 can not yield feasible solutions within 60 seconds are removed. The results show that, within 60 seconds of time limit, MaxFPS also outperforms SATLike3.0 on most classes of both PMS and WPMS instances. Specifically, for all the 32 (resp. 27) classes of PMS (resp. WPMS) instances, MaxFPS outperforms SATLike3.0 on 25 (resp. 20) classes, indicating again the excellent robustness of FPS that can boost SATLike3.0 in solving various classes of (W)PMS instances.


\begin{figure*}[!t]
\centering
\subfigure[Comparison on PMS\_2018]{
\includegraphics[width=0.7\columnwidth]{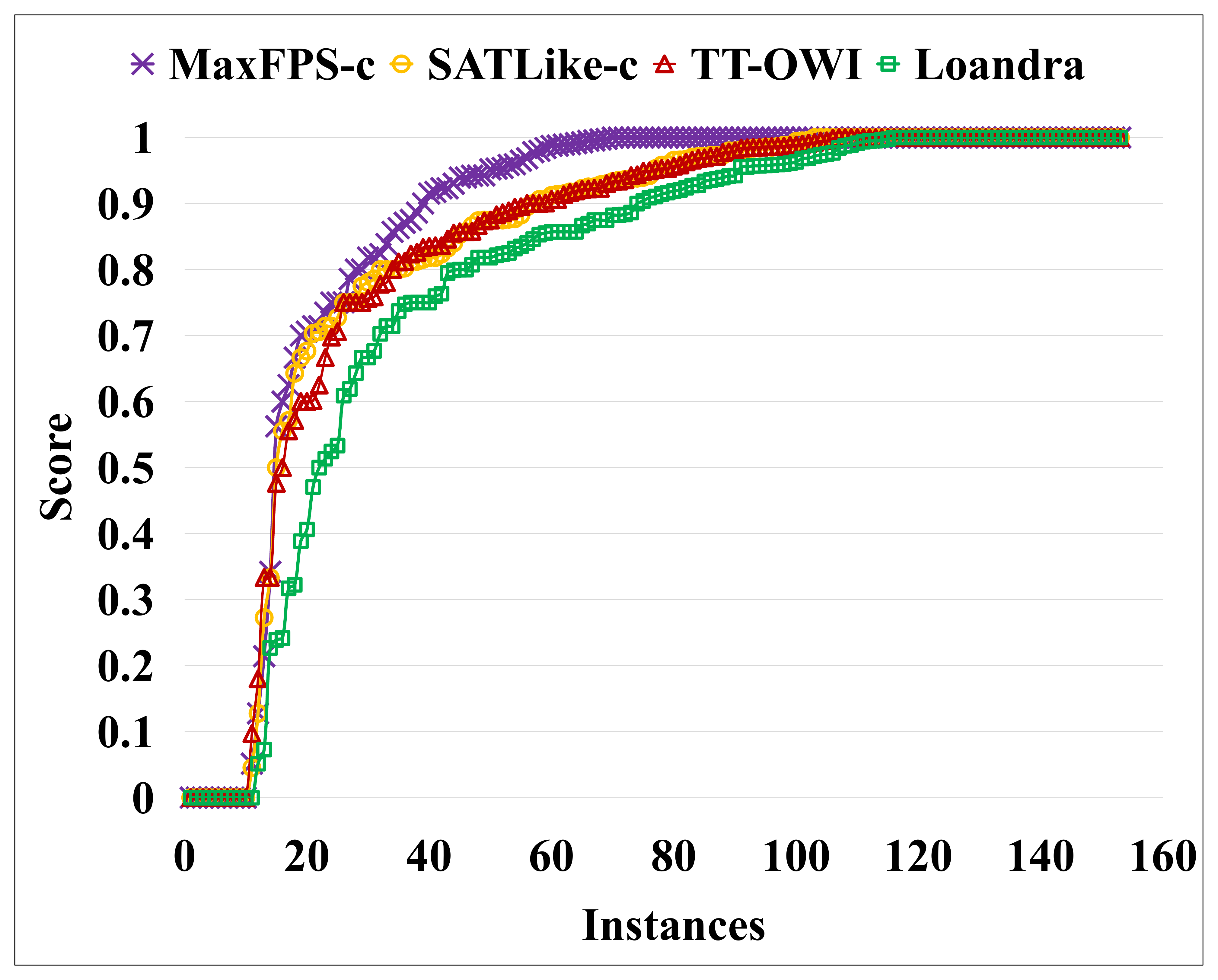}
\label{fig_PMS2018}}
\subfigure[Comparison on PMS\_2019]{
\includegraphics[width=0.7\columnwidth]{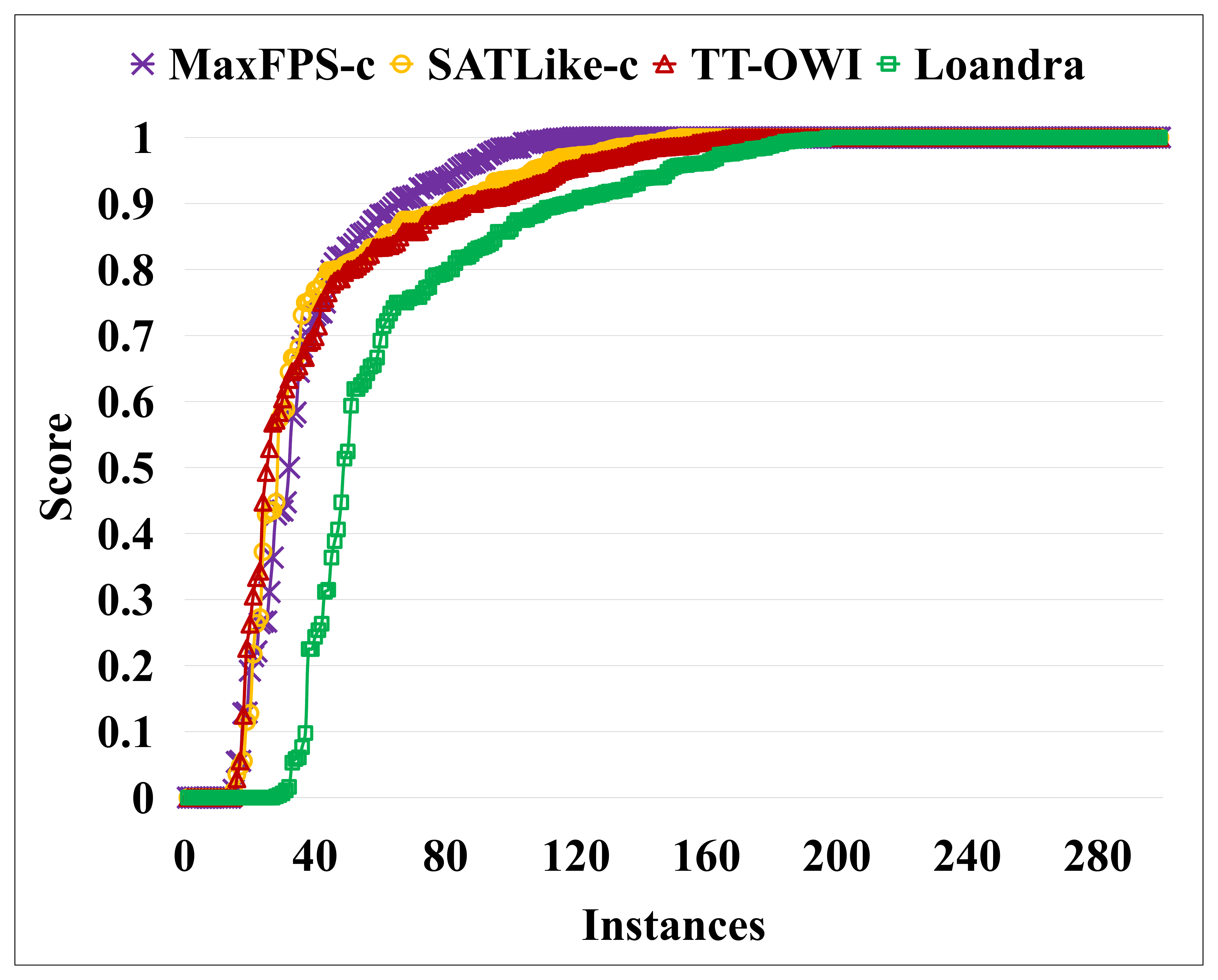}
\label{fig_PMS2019}}
\subfigure[Comparison on PMS\_2020]{
\includegraphics[width=0.7\columnwidth]{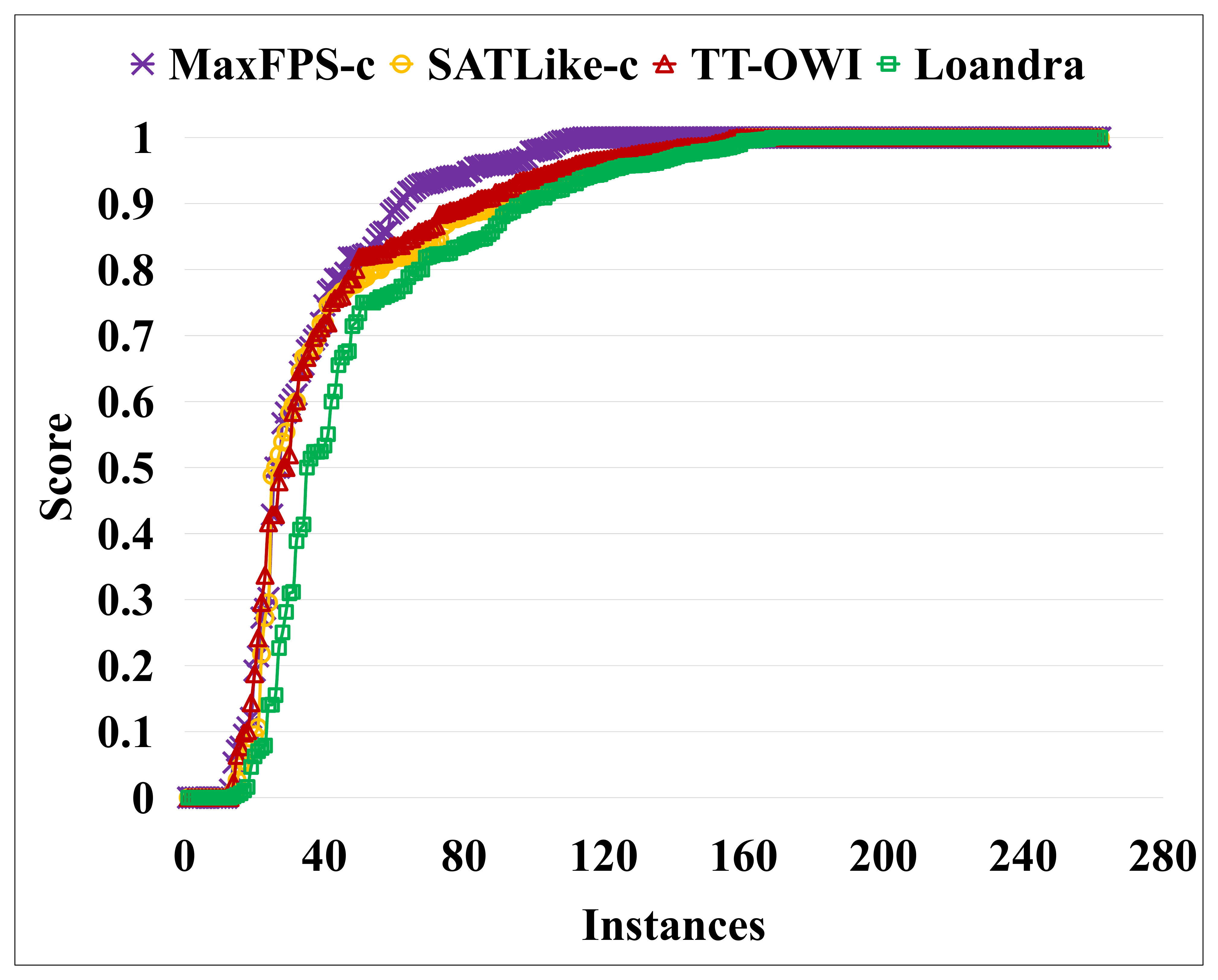}
\label{fig_PMS2020}}
\subfigure[Comparison on PMS\_2021]{
\includegraphics[width=0.7\columnwidth]{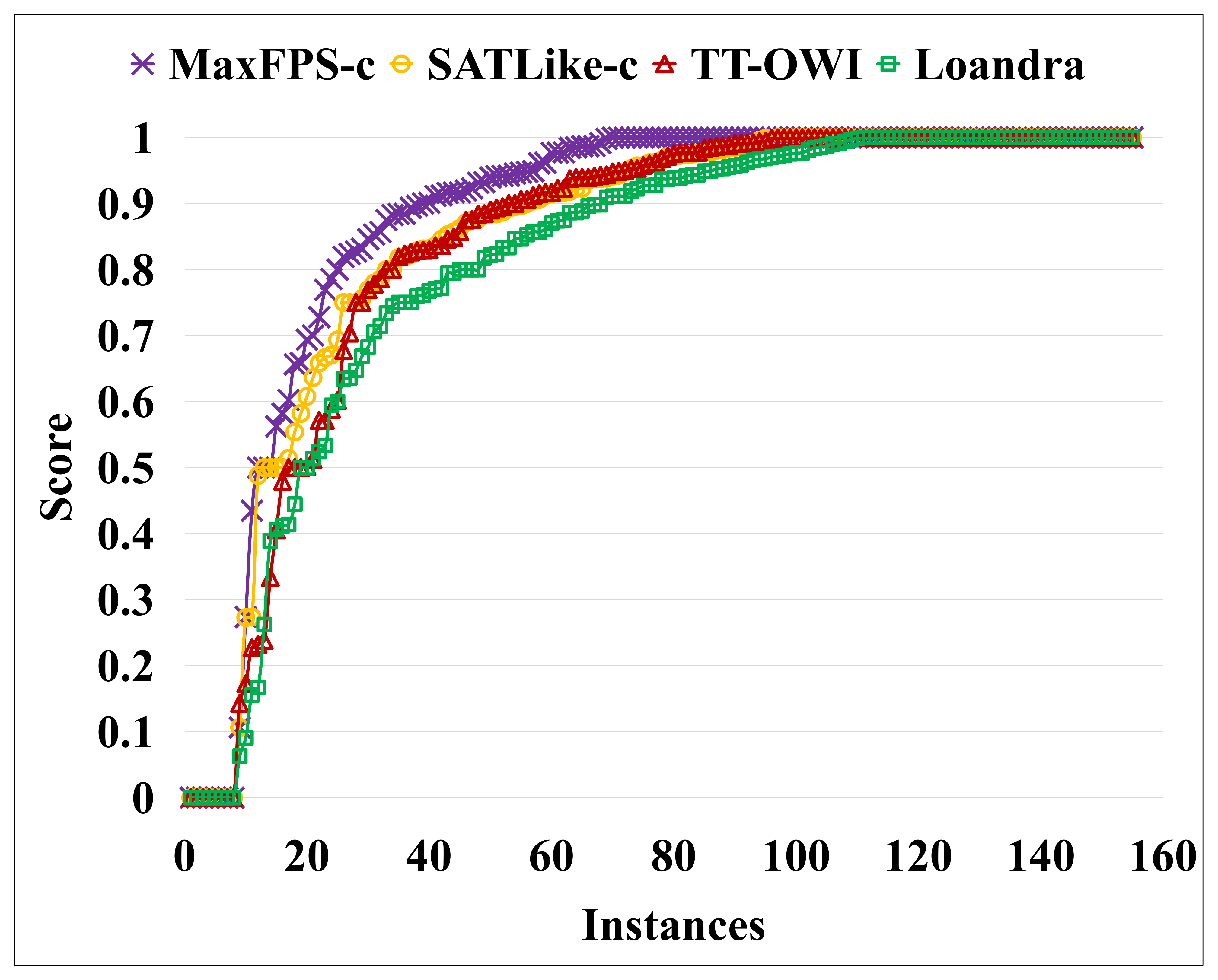}
\label{fig_PMS2021}}
\subfigure[Comparison on WPMS\_2018]{
\includegraphics[width=0.7\columnwidth]{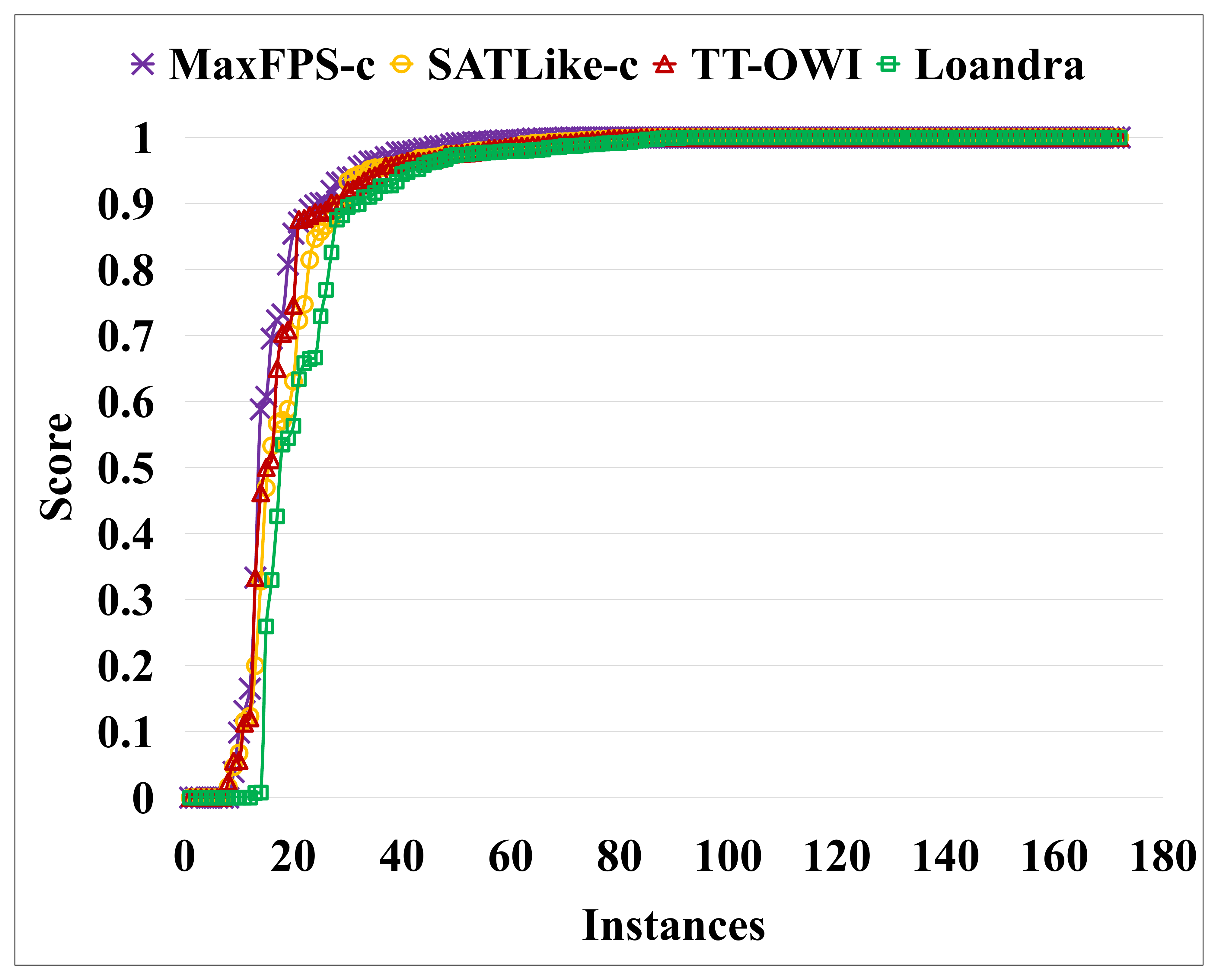}
\label{fig_WPMS2018}}
\subfigure[Comparison on WPMS\_2019]{
\includegraphics[width=0.7\columnwidth]{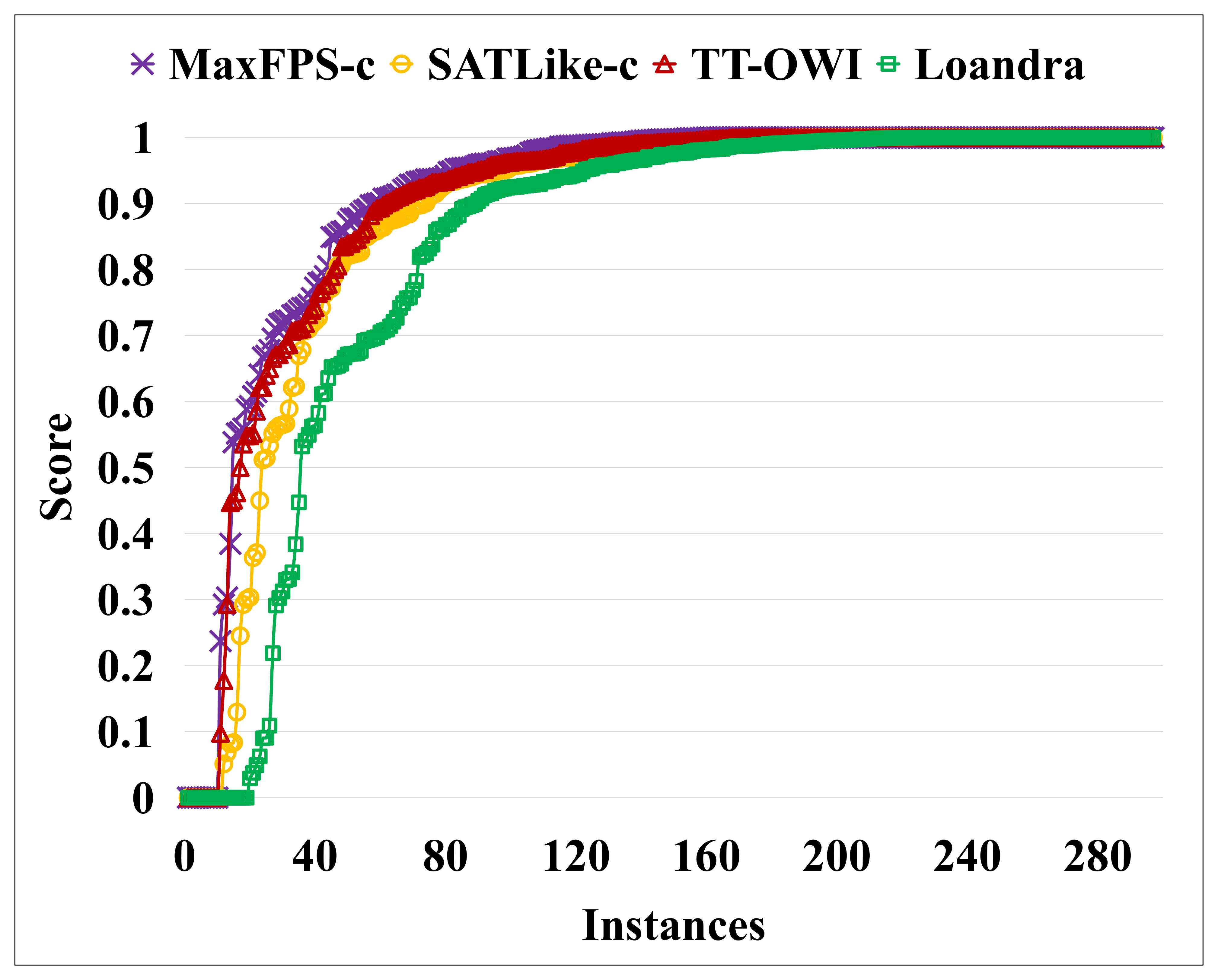}
\label{fig_WPMS2019}}
\subfigure[Comparison on WPMS\_2020]{
\includegraphics[width=0.7\columnwidth]{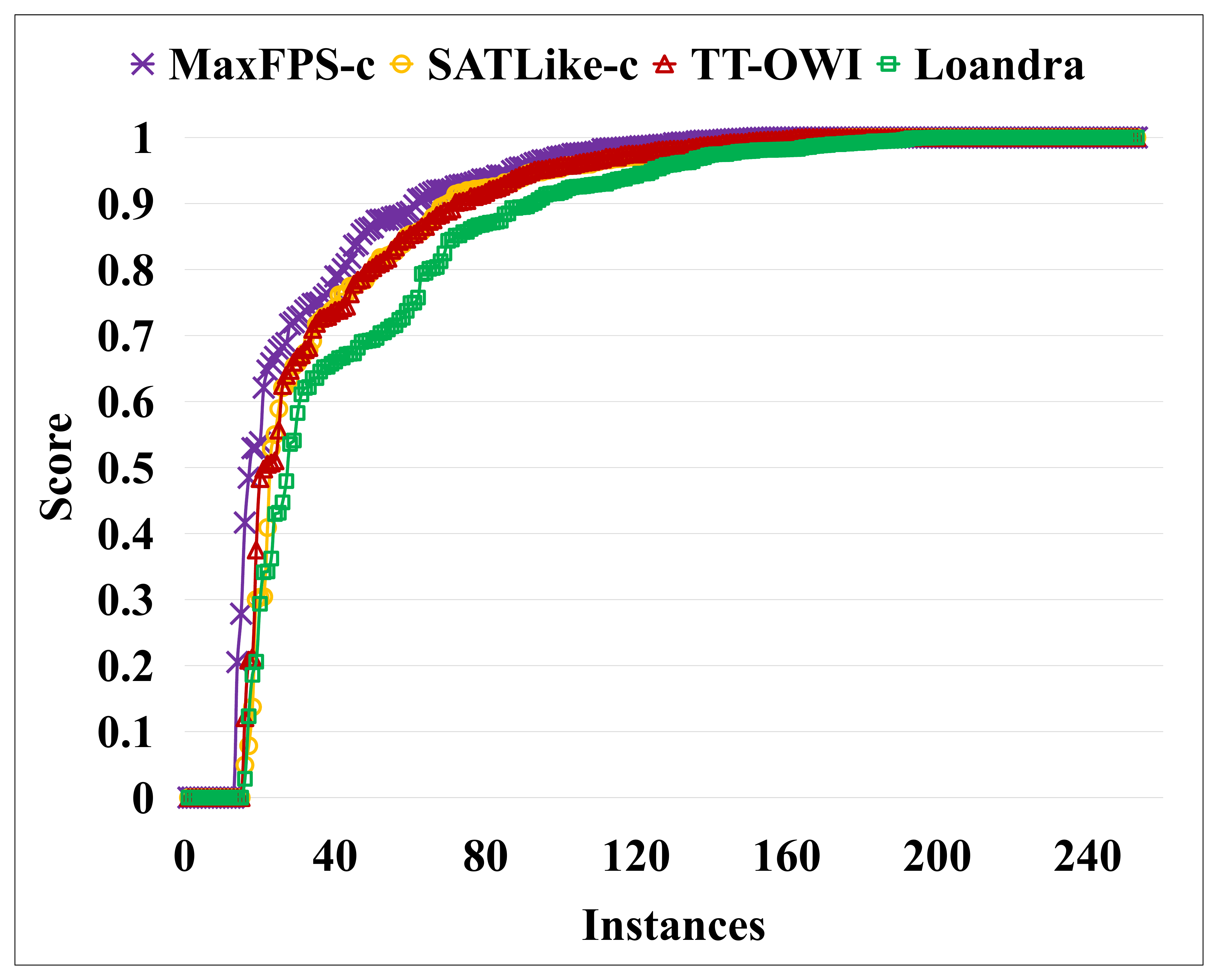}
\label{fig_WPMS2020}}
\subfigure[Comparison on WPMS\_2021]{
\includegraphics[width=0.7\columnwidth]{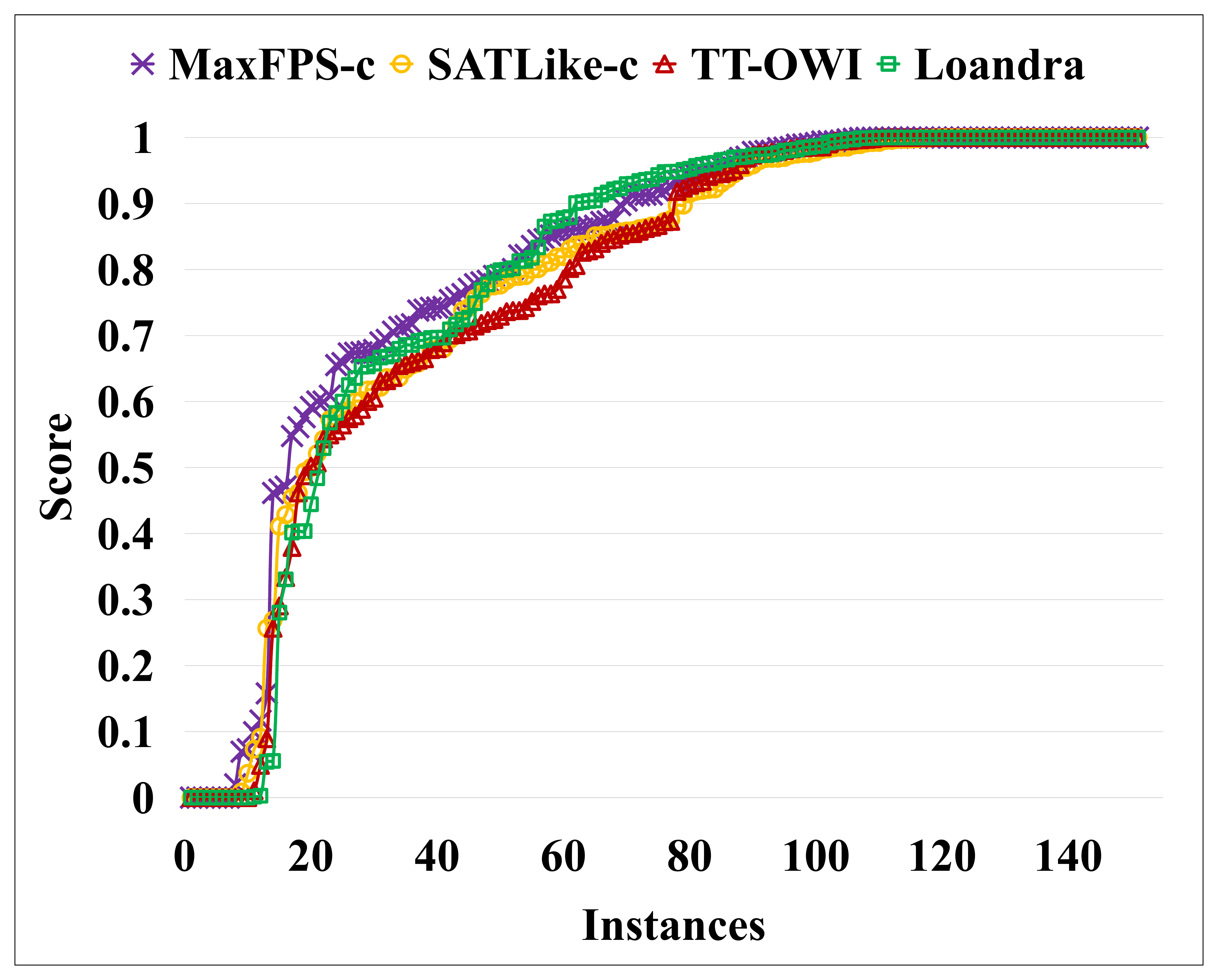}
\label{fig_WPMS2021}}
\caption{Distribution of scores per instance of MaxFPS-c, SATLike-c, TT-OWI, and Loandra on (W)PMS benchmarks.}
\label{fig_MaxFPS-c}
\end{figure*}

\section{Comparison with SAT-based Solvers}
We then present detailed comparison results of MaxFPS-c and the state-of-the-art SAT-based (W)PMS solvers, SATLike-c~\cite{Lei2021}, TT-Open-WBO-Inc (TT-OWI)~\cite{Nadel2019}, and Loandra~\cite{Berg2019}, within 300 seconds of time limit, in Figure \ref{fig_MaxFPS-c}. The results are expressed by the distributions of scores per instance of these SAT-based solvers as in MSEs\footnote{https://maxsat-evaluations.github.io/2019/results/incomplete/\\weighted-300s/summary.html}. To draw the results for each solver per benchmark, we first sort the scores obtained by the solver in solving the instances in the benchmark in ascending order, and then use a point to record each score and connect the points by a smooth curve.

The results in Figure \ref{fig_MaxFPS-c} show that the curves of MaxFPS-c are usually above the curves of the other three SAT-based solvers, indicating that in solving most of the instances, MaxFPS-c can yield better results.

\begin{table}[!t]
\footnotesize
\centering
\begin{tabular}{lrrrrrr} \bottomrule
\multirow{2}{*}{Benchmark} & \multirow{2}{*}{\#inst.} & \multicolumn{2}{c}{MaxFPS} &  & \multicolumn{2}{c}{MaxFPS$_4$} \\ \cline{3-4} \cline{6-7} 
                           &                          & \#win.          & time     &  & \#win.         & time         \\ \hline
PMS\_2018                  & 153                      & \textbf{106}    & 71.87    &  & 85             & 57.14        \\
PMS\_2019                  & 299                      & \textbf{198}    & 59.43    &  & 186            & 47.53        \\
PMS\_2020                  & 262                      & \textbf{177}    & 46.64    &  & 161            & 41.79        \\
PMS\_2021                  & 155                      & \textbf{96}     & 45.84    &  & 90             & 46.69        \\
WPMS\_2018                 & 172                      & \textbf{109}    & 60.61    &  & 92             & 59.30        \\
WPMS\_2019                 & 297                      & \textbf{188}    & 86.85    &  & 164            & 89.23        \\
WPMS\_2020                 & 253                      & \textbf{160}    & 85.17    &  & 133            & 91.40        \\
WPMS\_2021                 & 151                      & \textbf{78}     & 105.94   &  & 76             & 97.43    \\ \toprule  
\end{tabular}
\caption{Comparison of MaxFPS and MaxFPS$_4$.}
\label{table_V3}
\end{table}

\section{Additional Ablation Study}
Finally, we compare MaxFPS with its variant MaxFPS$_4$ to analyze the effectiveness of the early-stop strategy used in FPS (see Section 3.1). MaxFPS$_4$ is a variant of MaxFPS without early-stop strategy, \ie, MaxFPS$_4$ does not stop traversing the first-level variables when it finds a pair of variables that flipping both can improve the current solution. The comparison results of MaxFPS and MaxFPS$_4$ within 300 seconds of time limit are shown in Table \ref{table_V3}. From the results, we could observe that MaxFPS outperforms MaxFPS$_4$, indicating that the early-stop strategy in FPS is reasonable and effective, because it can improve the efficiency.

\end{document}